%% file: main.tex
\documentclass[twoside]{article}

\usepackage[accepted]{aistats2021}

\usepackage[utf8]{inputenc} 
\usepackage[T1]{fontenc}    
\usepackage{hyperref}       
\usepackage{url}            
\usepackage{booktabs}       
\usepackage{amsfonts}       
\usepackage{nicefrac}       
\usepackage{microtype}      

\usepackage{times}
\usepackage{graphicx}
\usepackage[ruled,vlined,noend]{algorithm2e}
\usepackage{amsmath}
\usepackage{amsthm}
\usepackage{indentfirst}
\usepackage{subcaption} 
\theoremstyle{definition}
\newtheorem{theorem}{Theorem}

\newtheorem{corollary}{Corollary}

\usepackage[perpage]{footmisc}  

\usepackage{datatool}
\usepackage{wrapfig}

\usepackage{chngcntr}
\usepackage{apptools}
\usepackage{float}
\AtAppendix{\counterwithin{prop}{section}}

\usepackage{authblk}
\author[1,2]{Yann Fraboni}
\author[2]{Richard Vidal}
\author[1]{Marco Lorenzi}
\affil[1]{University of C\^{o}te d’Azur, Inria, Epione Research Group, France}
\affil[2]{Accenture Labs, Sophia Antipolis, France}

\input{./tex/abbreviations}
\input{./tex/math_formulas}

\begin{document}

\twocolumn[

\aistatstitle{Free-rider Attacks on Model Aggregation in Federated Learning}

\aistatsauthor{ $\text{Yann Fraboni}^{1,2}$ \And $\text{Richard Vidal}^{2}$ \And  $\text{Marco Lorenzi}^{1}$ }

\aistatsaddress{ ${}^1$ Universit\'e C\^{o}te d’Azur, Inria Sophia Antipolis,
	Epione Research Group, France\\
	\and
	${}^2$ Accenture Labs, Sophia Antipolis, France\\ } ]

\input{./tex/Abstract}
\input{./tex/Introduction}
\input{./tex/convergence_analysis_alternate}

\input{./tex/Experiments}

\input{./tex/Conclusion}

\input{./tex/Acknowledgements}

\bibliographystyle{apalike}
\bibliography{main}

\input{./tex/Supplementary_Material_alternate}

\end{document}

%% file: tex/math_formulas.tex
\newcommand{\norm}[1]{\left\lVert#1\right\rVert}

\newcommand{\Dcal}[0]{\mathcal{D}}
\newcommand{\Lcal}[0]{\mathcal{L}}

\newcommand{\expect}{\operatorname{\mathbb{E}}}
\newcommand{\Var}{\operatorname{Var}}
\newcommand{\Cov}{\operatorname{Cov}}
\newcommand{\VAR}[1]{\Var\left[#1\right]}
\newcommand{\COV}[1]{\Cov\left[#1\right]}
\newcommand{\E}[1]{\expect\left[#1\right]}
\newcommand{\EE}[2]{\expect_{#1}\left[#2\right]}



%% file: tex/Abstract.tex
\begin{abstract}

Free-rider attacks against federated learning  consist in dissimulating participation to the federated learning process with the goal of obtaining the final aggregated model without actually contributing with any data. This kind of attacks is critical in sensitive applications of federated learning, where data is scarce and the model has high commercial value. We introduce here the first theoretical and experimental analysis of free-rider attacks on federated learning schemes based on iterative parameters aggregation, such as FedAvg or FedProx, and provide formal guarantees for these attacks to converge to the aggregated models of the fair participants. We first show that a straightforward implementation of this attack can be simply achieved by not updating the local parameters during the iterative federated optimization. As this attack can be detected by adopting simple countermeasures at the server level, we subsequently study more complex disguising schemes based on stochastic updates of the free-rider parameters. We demonstrate the proposed strategies on a number of experimental scenarios, in both iid and non-iid settings. We conclude by providing recommendations to avoid free-rider attacks in real world applications of federated learning, especially in sensitive domains where security of data and models is critical.

\end{abstract}

%% file: tex/Introduction.tex
\section{Introduction}

Federated learning is a training paradigm that has gained popularity in the last years as it enables different clients to jointly learn a global model without sharing their respective data. It is particularly suited for Machine Learning applications in domains where data security is critical, such as healthcare \cite{Brisimi,silva2019federated}. The relevance of this approach is witnessed by current large scale federated learning initiatives under development in the medical domain, for instance for learning predictive models of breast cancer\footnote{\url{blogs.nvidia.com/blog/2020/04/15/federated-learning-mammogram-assessment/}}, or for drug discovery and development\footnote{\url{www.imi.europa.eu/projects-results/project-factsheets/melloddy}}.

The participation to this kind of research initiatives is usually exclusive and typical of applications where data is scarce and unique in its kind. In these settings, aggregation results entail critical information beyond data itself, since a model trained on exclusive datasets may have very high commercial or intellectual value. For this reason, providers may not be interested in sharing the model: the commercialization of machine learning products  would rather imply the availability of the model as a service through web- or cloud-based API. This is due to the need of preserving the intellectual property on the model components, as well as to avoid potential information leakage, for example by limiting the maximum number of queries allowed to the users \cite{Carlini, Fredrikson-MI-2015, Ateniese2015}.  

This critical aspect can lead to the emergence of opportunistic behaviors in federated learning, where ill-intentioned clients may participate with the aim of obtaining the federated model, without actually contributing with any data during the training process.  In particular, the attacker, or free-rider, aims at disguising its participation to federated learning while ensuring that the iterative training process ultimately converges to the wished target: the aggregated model of the fair participants. Free-riding attacks performed by ill-intentioned participants ultimately open federated learning initiatives to intellectual property loss and data privacy breaches, taking place for example in the form of model inversion  \cite{Fredrikson-MI-2014,Fredrikson-MI-2015}. 

The study of security and safety of federated learning is an active research domain, and several kind of attacks are matter of ongoing studies. For example, an attacker may interfere during the iterative federated learning procedure to degrade/modify models performances \cite{Bhagoji2019,LiDataPoisoning2016,Yin2018,xie2019dba,Shen2016}, or retrieve information about other clients' data \cite{Wang2019,Hitaj2017}.
{Since currently available defence methods such as \cite{Fung2020, Bhagoji2019} are generally based on outliers detection mechanisms, they are generally not suitable to prevent free-riding, as this kind of attack is explicitly conceived to stay undetected while not perturbing the FL process. }
Free-riding may become a critical aspect of future machine learning applications, as federated learning is rapidly emerging as the standard training scheme in current cooperative learning initiatives. To the best of our knowledge, the only investigation is in a preliminary work \cite{Lin2019} focusing on attack strategies operated on federated learning based on gradient aggregation. However, no theoretical guarantees are provided for the effectiveness of this kind of attacks. Furthermore this setup is unpractical in many real world applications, where federated training schemes based on model averaging are instead more common, due to the reduced data exchange across the network. FedAvg \cite{BrendanMcMahan2017} is the most representative framework of this kind, as it is based on the iterative averaging of the clients models’ parameters, after updating each client model for a given number of training epochs at the local level. To improve the robustness of FedAvg in non-iid and heterogeneous learning scenarios, FedProx \cite{LiFedProx2018} extends FedAvg by including a regularization term penalizing local departures of clients' parameters from the global model.

%

The contribution of this work consists in the development of a theoretical framework for the study of free-rider attacks in federated learning schemes based on model averaging, such as in FedAvg and FedProx. The problem is here formalized via the reformulation of federated learning as a stochastic process describing the evolution of the aggregated parameters across iterations. To this end, we build upon previous works characterizing the evolution of model parameters in 
Stochastic Gradient Descent (SGD) as a continuous time process \cite{Mandt2017,Orvieto2018,Li2017,He2018}. A critical requirement for opportunistic free-rider attacks is to ensure the convergence of the training process to the wished target represented by the aggregated model of the fair clients. We show that the proposed framework allows to derive explicit conditions to guarantee the success of the attack. This is an important theoretical feature as it is of primary interest for the attacker  to not interfere with the learning process. 

We first derive in Section \ref{plain_free-rider} a basic free-riding strategy to guarantee the convergence of federated learning to the model of the fair participants. This strategy simply consists in returning at each iteration the received global parameters. As this behavior could easily be detected by the server, we build more complex strategies to disguise the free-rider contribution to the optimization process, based on opportune stochastic perturbations of the parameters. We demonstrate in Section \ref{disguised_free-rider} that this strategy does not alter the global model convergence, and in Section \ref{experiments} we experimentally demonstrate our theory on a number of learning scenarios in both iid and non-iid settings. All proofs and additional material are provided in the Appendix.



%% file: tex/convergence_analysis_alternate.tex
\section{Methods}

Before introducing in Section \ref{free-rider_strategies} the core idea of free-rider attacks, we first recapitulate in Section \ref{recap_FL} the general context of parameter aggregation in federated learning.

\subsection{Federated learning through model aggregation: FedAvg and FedProx}
\label{recap_FL}

In federated learning, we consider a set $I$ of participating clients respectively owning  datasets $\Dcal_i$ composed of  $M_i$ samples. During optimization, it is generally assumed that the $D$ elements of the clients' parameters vector $\boldsymbol{{\theta}}_i^t=({\theta}_{i,0}^t,{\theta}_{i,1}^t,...,{\theta}_{i,D}^t),$ and the global parameters ${\boldsymbol\theta}^t=(\theta_{0}^t,\theta_{1}^t,...,\theta_{D}^t)$  are aggregated independently at each iteration round $t$. Following this assumption, and for simplicity of notation, in what follows we restrict our analysis to a single parameter entry, that will be generally denoted  by ${\theta}_i^t$ and $\theta^t$ for  clients and server respectively.

In this setting, to estimate a global model across clients, FedAvg \cite{BrendanMcMahan2017} is an iterative training strategy based on the aggregation of local model  parameters ${\theta}_i^t$. At each iteration step $t$, the server sends the current global model parameters $\theta^t$ to the clients. Each client updates the model by minimizing over $E$ epochs the local cost function $\Lcal({\theta}_i^{t+1},\Dcal_i)$ initialized with $\theta^t$, and subsequently returns the updated local parameters ${\theta}_i^{t+1}$ to the server. The global model parameters $\theta^{t+1}$ at the iteration step $t+1$ are then estimated as a weighted average:
\begin{equation}
\label{FedAvg_server_aggregation}
\theta^{t+1}=\sum_{i\in I}\frac{M_i}{N}{\theta}_i^{t+1},
\end{equation}
where $N=\sum_{i\in I}M_i$ represents the total number of samples across distributed datasets. FedProx \cite{LiFedProx2018} builds upon FedAvg by adding to the cost function a L2 regularization term penalizing the deviation of the local parameters ${\theta}_i^{t+1}$ from the global parameters $\theta^t$. The new cost function is \smash{$\Lcal_{Prox}({\theta}_i^{t+1},\Dcal_i,\theta^t)=\Lcal({\theta}_i^{t+1},\Dcal_i)+\frac{\mu}{2}\norm{{\theta_i}^{t+1}-\theta^t}^2$} where $\mu$ is the hyperparameter monitoring the regularization by enforcing proximity between local update ${\theta_i}^{t+1}$ and reference model $\theta^t$.

\subsection{Formalizing Free-rider attacks}
\label{free-rider_strategies}

\begin{algorithm}
	\SetAlgoLined
	\KwIn{learning rate $\lambda$, epochs $E$, initial model $\theta^0$, batch size $S$}
	$\tilde{\theta}^0 = \theta^0$;\\
	\For{each round t=0,...,T-1}{
		
		Send the global model $\tilde{\theta}^t$ to all the clients;
		
		\For{each fair client $j\in J$}{
			$\tilde{\theta}_j^{t+1}=ClientUpdate(\tilde{\theta}^t,E,\lambda)$;
			
			Send $\tilde{\theta}_j^{t+1}$ to the server;}
		
		\For{each free-rider $k\in K$}{
			
			\eIf{disguised free-rider}{$\tilde{\theta}_{k}^{t+1} = \tilde{\theta}^t+\epsilon$, where $\epsilon \sim \mathcal{N}(0 , \sigma_k^2 ) $;}{$ \tilde{\theta}_k^{t+1} = \tilde{\theta}^t$}
			
			Send $\tilde{\theta}_k^{t+1}$ to the server;}
		
		$\tilde{\theta}^{t+1}=\sum_{j\in J}\frac{M_j}{N} \tilde{\theta}_j^{t+1}+\sum_{k\in K}\frac{M_k}{N} \tilde{\theta}_k^{t+1}$;
			
	}
\caption{Free-riding in federated learning}\label{algo:freeriding}
\end{algorithm}

Aiming at obtaining the aggregated model of the fair clients, the strategy of a free-rider consists in participating to federated learning by dissimulating local updating through the sharing of opportune counterfeited parameters. The free-riding attacks investigated in this work are illustrated in Algorithm \ref{algo:freeriding}, and analysed in the following sections from both theoretical and experimental standpoints.

We denote by $J$ the set of fair clients, i.e. clients following the federated learning strategy of Section \ref{recap_FL} and by $K$ the set of free-riders, i.e.
malicious clients pretending to participate to the learning process, such that $I=J\cup K$ and $J\neq \emptyset$. We denote by $M_K$ the  number of samples declared by the free-riders.

\subsection{SGD perturbation of the fair clients local model}
\label{sec:sgd_perturbation}
To describe the clients' parameters observed during federated learning, we rely on the modeling of Stochastic Gradient Descent (SGD) as a continuous time stochastic process \cite{Mandt2017,Orvieto2018,Li2017,He2018}.

For a client $j$, let us consider the following form for the loss function:
\begin{equation}
\label{eq:full_loss_function}
\mathcal{L}_j(\theta_j)=\frac{1}{M_j}\sum_{n=1}^{M_j}l_{n,j}(\theta_j),\ 
\end{equation}
where $M_j$ is the number of samples owned by the client, and $l_{n,j}$ is the contribution to the overall loss from a single observation $\{x_{n,j};y_{n,j}\}$. The gradient of the loss function is defined as $g_j(\theta_j)\equiv\nabla\mathcal{L}_j(\theta_j)$.

We represent SGD by considering a minibatch $\mathcal{S}_{j,k}$, composed of a set of $S$ different indices drawn uniformly at random from the set $\{1,\ ...\ ,M_j\}$, each of them indexing a function $l_{n,j}(\theta_j)$ and where $k$ is the index of the minibatch. Based on $\mathcal{S}_{j,k}$, we form a stochastic estimate of the loss,
\begin{equation}
\label{eq:stochastic_loss_function}
\mathcal{L}_{\mathcal{S}_{j,k}}(\theta_j)=\frac{1}{S}\sum_{n\in \mathcal{S}_{j,k}}^{}l_{n,j}(\theta_j),\ 
\end{equation}
where the corresponding stochastic gradient  is defined as $g_{\mathcal{S}_{j,k}}(\theta_j)\equiv\nabla\mathcal{L}_{\mathcal{S}_{j,k}}(\theta_j)$.

By observing that gradient descent is a sum of $S$ independent and uniformly distributed samples, thanks to the central limit theorem, gradients at the client level can thus be modeled by a Gaussian distribution 
\begin{equation}
\label{eq:gradient_gaussian}
g_{\mathcal{S}_{j,k}}(\theta_j)
\sim \mathcal{N}(g_j(\theta_j),\frac{1}{S}\sigma_j^2(\theta_j)),
\end{equation}
where $g_j(\theta_j)=\EE{s}{g_{\mathcal{S}_{j,k}}(\theta_j)}$ is the full gradient of the loss function in equation (\ref{eq:full_loss_function}) and $\sigma_j^2(\theta_j)$ is the variance associated with the loss function in equation (\ref{eq:stochastic_loss_function}).

SGD updates are expressed as:
\begin{equation}
\label{eq:gradient_local_update}
\theta_j(u_j+1)=\theta_j(u_j)-\lambda g_{\mathcal{S}_{j,k}}(\theta_j(u_j)),
\end{equation}
where $u_j$ is the SGD iteration index and $\lambda$ is the learning rate set by the server.

By defining $\Delta \theta_j(u_j)=\theta_j(u_j+1)-\theta_j(u_j)$, we can rewrite the update process as
\begin{equation}
\Delta\theta_j(u_j)=-\lambda g_j(\theta_j(u_j))+\frac{\lambda}{\sqrt{S}}\sigma_j(\theta_j)\Delta W_j,
\end{equation}
where $ \Delta W_j\sim \mathcal{N}(0,1)$. The resulting continuous-time model \cite{Mandt2017,Orvieto2018,Li2017,He2018} is
\begin{equation}
\label{eq:SDE_SGD}
\mathrm{d}\theta_j=-\lambda g_j(\theta_j)\mathrm{d}u_j+\frac{\lambda}{\sqrt{S}}\sigma_j(\theta_j)\mathrm{d}W_j.
\end{equation}
where $W_j$ is a continuous time Wiener Process.

Similarly as in \cite{Mandt2017}, we assume that $\sigma_j(\theta_j)$ is approximately constant with respect to $\theta_j$ for the client's stochastic gradient updates between $t$ and $t+1$, and will therefore denote $\sigma_j(\theta_j)=\sigma_j^t$. Following \cite{Mandt2017}, we consider a local quadratic approximation for the client's loss, leading to a linear form for the gradient $g_j(\theta_j)\simeq r_j [ \theta_j - \theta_j^* ]$, where $r_j\in \mathbb{R}^+$ depends on the approximation of the cost function around the local minimum $\theta_j^*$. This assumption enables rewriting equation (\ref{eq:SDE_SGD}) as an Ornstein-Uhlenbeck process \cite{Ornstein}. Starting from the initial condition represented by $\theta^t$, the global model received at the iteration $t$, we characterize the local updating of the parameters through equation (\ref{eq:SDE_SGD}), and we follow the evolution up to the time $\frac{EM_j}{S}$, where $E$ is the number of epochs, and $M_j$ is the number of samples owned by the client. Assuming that $M_j$ is a multiple of $S$, the number of samples per minibatch, the quantity $\frac{EM_j}{S}$ represents the total number of SGD steps run by the client. The updated model ${\theta}_j^{t+1}$ uploaded to the server therefore takes the form:
\begin{align}
\label{eq:OU_close_form}
{\theta}_j^{t+1}
&=\underbrace{e^{-\lambda r_j\frac{EM_j}{S}}[\theta^t-\theta_j^*]+\theta_j^*}_{\hat{\theta}_j^{t+1}}\nonumber\\
&+\frac{\lambda}{\sqrt{S}}\int_{u=0}^{\frac{EM_j}{S}}e^{-\lambda r_j\left(\frac{EM_j}{S}-u\right)}\sigma_j^tdW_u.
\end{align}

We note that the relative number of SGD updates for the fair clients, $\frac{EM_j}{S}$, influences the parameter $\eta_j=e^{-\lambda r_j\frac{EM_j}{S}}$, which becomes negligible for large values of $E$. 

The variance introduced by SGD can be rewritten as 
\begin{align}
\VAR{{\theta}_j^{t+1}|\theta^t}
&=\underbrace{\frac{\lambda}{S}{\sigma_j^t}^2\frac{1}{2r_j}\left[1-e^{-2\lambda r_j\frac{EM_j}{S}}\right]}_{{\rho_j^t}^2},
\label{eq:variance_SGD}
\end{align}
where we can see that the higher $\frac{EM_j}{S}$, the lower the overall SGD noise. The noise depends on the local loss function $r_j$, on the server parameters (number of epochs $E$, learning rate $\lambda$, and number of samples per minibatch $S$), and on the clients' data specific parameters (SGD variance ${\sigma_j^t}^2$ ). 

Equation (\ref{eq:OU_close_form}) shows that clients' parameters observed during federated learning can be expressed as ${\theta}_j^t = \hat{\theta}_j^t + \rho_j^t \zeta_{j,t}$, where, given $\theta^t$, $\hat{\theta}_j^t$ is a deterministic component corresponding to the model obtained with $\frac{EM_j}{S}$ steps of gradient descents, and $\zeta_{j,t}$ is a delta-correlated Gaussian white noise. We consider in what follows a constant local noise variance $\sigma_j^2$ (this assumption will be relaxed in Section \ref{noise_fair_clients} to consider instead time-varying noise functions $\rho_j^t$). 

Based on this formalism, in the next Section we study a basic free-rider strategy simply consisting in returning at each iteration the received global parameters.
We call this type of attack \textit{plain free-riding}.

\subsection{Plain free-riding}
\label{plain_free-rider}

We denote by $\tilde{\theta}$ and $\tilde{\theta}_j$ respectively the global and local model parameters obtained in presence of free-riders. The plain free-rider returns the same model parameters as the received ones, i.e. $\forall k\in K,\ \tilde{\theta}_k^{t+1}= \tilde{\theta}^t$. In this setting, the server aggregation process (\ref{FedAvg_server_aggregation}) can be rewritten as:
\begin{equation}
\label{neutral_server_aggregation}
\tilde{\theta}^{t+1}
=\sum_{j\in J}\frac{M_j}{N}\tilde{\theta}_j^{t+1}+\frac{M_K}{N}\tilde{\theta}^t\ ,
\end{equation}
where $\tilde{\theta}^t$ is the global model and $\tilde{\theta}_j^t$ are the fair clients' local models uploaded to the server for free-riding. 

\subsubsection{Free-riders perturbation of the fair clients local model}
In this section, we investigate the effect of the free-riders on the local optimization performed by the fair clients at every server iteration. The participation of the free-riders to federated learning implies that the processes of the fair clients are being perturbed by the attacks throughout training. In particular, the initial conditions of the local optimization problems are modified according to the perturbed aggregation of equation (\ref{neutral_server_aggregation}).

Back to the assumptions of Section \ref{sec:sgd_perturbation} , the initial condition $\tilde{\theta}^t$ of the local optimization includes now the aggregated model of the fair clients and a perturbation coming from the free-riders. Thus, equation (\ref{eq:OU_close_form}) in presence of free-riding can be written as 
\begin{align}
\label{eq:OU_close_form_rewritten_for_fr}
\tilde{\theta}_j^{t+1}
&=\eta_j[\tilde{\theta}^t-\theta_j^*]+\theta_j^*\nonumber\\
&+\frac{\lambda}{\sqrt{S}}\int_{u=0}^{\frac{EM_j}{S}}e^{-\lambda r_j\left(\frac{EM_j}{S}-u\right)}\tilde{\sigma}_j^t dW_u,
\end{align}
where  $\tilde{\sigma}_j^t=\sigma_j^t(\tilde{\theta}_j)$ is the SGD variance for free-riding. We consider that $\tilde{\sigma}_j^t=\sigma_j^t=\sigma_j$. This assumption will be relaxed in Section
\ref{noise_fair_clients} to consider instead time-varying noise functions. With analogous considerations to those made in Section \ref{sec:sgd_perturbation}, the updated parameters take the form:
\begin{align}
\tilde{\theta}_j^{t+1}
&=\eta_j[\tilde{\theta}^t-\theta_j^*]+\theta_j^*+\rho_j\tilde{\zeta}_{j,t},
\end{align}
where $\tilde{\zeta}_{j,t}$ is a delta-correlated Gaussian white noise. Similarly as for federated learning, $\smash{\E{\tilde{\theta}_j^{t+1}|\tilde{\theta}^t}}=\eta_j[\tilde{\theta}^t-\theta_j^*]+\theta_j^*$, and \smash{$\VAR{\tilde{\theta}_j^{t+1}|\tilde{\theta}^t}=\rho_j^2$}.




We want to express the global optimization process $\tilde{\theta}^t$ due to free-riders in terms of a a perturbation of the equivalent stochastic process $\theta^t$ obtained with fair clients only. Theorem \ref{theo:diff_plain_fr} provides a recurrent form for the difference between these two processes. 


\begin{theorem}[]\label{theo:diff_plain_fr} Under the assumptions of Section  \ref{sec:sgd_perturbation} and \ref{plain_free-rider} for the local optimization processes resulting from federated learning with respectively only fair clients and with free-riders, the difference between the aggregation processes of formulas (\ref{FedAvg_server_aggregation}) and (\ref{neutral_server_aggregation}) takes the following recurrent form: 
    \begin{align}\label{eq:recurrent}
    \tilde{\theta}^t - \theta^t
    & = \sum_{i=0}^{t-1}\left( \epsilon + \frac{M_K}{N} \right)^{t - i -1} f(\theta^i)\\\nonumber
    &+ \sum_{i=0}^{t-1}\left( \epsilon + \frac{M_K}{N} \right)^{t - i -1} ( \tilde{\nu}_i - \nu_i),
    \end{align}
    with $f(\theta^t) 
    = \frac{M_K}{N} \left[\theta^t - \sum_{j\in J} \frac{M_j}{N - M_K} [\eta_j (\theta^t -\theta_j^*) + \theta_j^*]\right]$,
    $\epsilon = \sum_{j\in J} \frac{ M_j}{ N } \eta_j$, $\nu_t = \sum_{j\in J} \frac{ M_j}{ N - M_K} \rho_j \zeta_{j, t}$
    and $\tilde{\nu}_t = \sum_{j\in J} \frac{ M_j}{ N } \rho_j \tilde{\zeta}_{j, t}$.
\end{theorem}

We note that in the special case with no free-riders (i.e. $M_K=0$), the quantity $\tilde{\theta}^t - \theta^t$ depends on the second term  of equation (\ref{eq:recurrent}) only, and represents the comparison between two different realizations of the stochastic process associated to the federated global model. Theorem \ref{theo:diff_plain_fr} shows that in this case the variance across optimization results is non-zero, and depends on the intrinsic variability of the local optimization processes quantified by the variable $\nu_t$. We also note that in presence of free-riders the convergence to the model obtained with fair clients depends on the relative sample size declared by the free-riders $\frac{M_K}{N}$.

\subsubsection{Convergence analysis of plain free-riding}

Based on the relationship between the learning processes established in Theorem \ref{theo:diff_plain_fr},  we are now able to prove that federated learning with plain free-riders defined in equation (\ref{neutral_server_aggregation}) converges in expectation to the aggregated model of the fair clients of equation (\ref{FedAvg_server_aggregation}). 

\begin{theorem}[Plain free-riding] \label{deterministic_theorem} Assuming FedAvg converges in expectation, and based on the assumption of Theorem \ref{theo:diff_plain_fr},  the following asymptotic properties hold:
	\begin{align}
	    \E{\tilde{\theta}^t - \theta^t}
	    & \xrightarrow{ t\rightarrow+\infty} 0,\\
	    \VAR{\tilde{\theta}^t - \theta^t}
	    &\xrightarrow{t\rightarrow+\infty} \frac{ [\frac{1}{N^2} + \frac{1}{(N-M_K)^2}] \sum_{j\in J} \left( M_j\rho_j\right)^2 }{1- \left( \epsilon + \frac{M_K}{N} \right)^{2}}
.\label{plain:variance}
	\end{align}  
\end{theorem}

As a corollary of Theorem \ref{deterministic_theorem}, in Proof \ref{deterministic_proof} it is shown that the asymptotic variance is strictly increasing with the sample size $M_K$ declared by the free-riders.  In practice, the smaller the total number of data points declared by the free-riders, the closer the final aggregation result approaches the model obtained with fair clients only. On the contrary, when the the sample size of the fair clients is negligible with respect to the the one declared by the free-riders, i.e. $N \simeq M_K$, 
the variance tends to infinity. This is due to the ratio approaching to 1 in the geometric sum of the second term of equation (\ref{eq:recurrent}). 
In the limit case when only free-riders participate to federated learning ($J =\emptyset $), we obtain instead the trivial result $\tilde{\theta}^t = \theta^0$ and $\VAR{\tilde{\theta}^t}=0$. 
In this case there is no learning throughout the training process. Finally, with no free-riders ($M_K=0$), we obtain $\VAR{\tilde{\theta}_1^t - \theta_2^t}\xrightarrow{t\rightarrow+\infty} \frac{2}{N^2}  \frac{1}{1- \epsilon  ^{2}}\sum_{j\in J} \left( M_j\rho_j\right)^2$, reflecting the variability of the fair aggregation process due to the stochasticity of the local optimization processes.

\subsection{Disguised free-riding}
\label{disguised_free-rider}

Plain free-riders can be easily detected by the server, since for each iteration the condition $[ \tilde{\theta}_k^{t+1}-\tilde{\theta}^t=0]$ is true. 
In what follows, we study improved attack strategies based on the sharing of opportunely disguised parameters, and investigate sufficient conditions on the disguising models to obtain the desired convergence behavior of free-rider attacks.

\subsubsection{Additive noise to mimic SGD updates}
A disguised free-rider with additive noise generalizes the plain one, and uploads parameters $\tilde{\theta}_k^{t+1}= \tilde{\theta}^t + \varphi_k(t) \epsilon_t$. Here, the perturbation $\epsilon_{t}$  is assumed to be Gaussian white noise, and $\varphi_k(t)>0$ is a suitable time-varying perturbation compatible with the free-rider attack. 
As shown in equation (\ref{eq:OU_close_form}), the parameters uploaded by the fair clients take the general form composed of an expected model corrupted by a stochastic perturbation due to SGD. Free-riders can mimic this update form by adopting a noise structure similar to the one of the fair clients:
\begin{align}
\label{eq:optimal_varphi_general}
\varphi_k^2(t)
&=\frac{\lambda}{S}{\sigma_k^t}^2\frac{1}{2r_k}\left[1-e^{-2\lambda r_k\frac{EM_k}{S}}\right],
\end{align} 
where $r_k$ and $\sigma_k^t$ would ideally depend on the (non-existing) free-rider data distribution and thus need to be determined, while $M_k$ is the declared number of samples.
Compatibly with the assumptions of constant SGD variance $\sigma_j^2$ for the fair clients, we here assume that the free-riders noise is constant and compatible with the SGD form:
%
\begin{align}
\label{eq:optimal_varphi_cte}
\varphi_k^2
&=\frac{\lambda}{S}\sigma_k^2\frac{1}{2r_k}\left[1-e^{-2\lambda r_k\frac{EM_k}{S}}\right].
\end{align}

{The parameters $r_k$ and $\sigma_k$ affect the noise level and decay of the update, and thus the ability of the free-rider of mimicking a realistic client. These parameters can be ideally estimated by computing a plausible quadratic approximation of the local loss function (Section \ref{sec:sgd_perturbation}). While the estimation may require the availability of some form of data for the free-rider, in Section \ref{sec:attacks_disguised_convergence} we prove that, for any combination of $r_k$ and $\sigma_k$, federated learning still converges to the desired aggregated target.}

Analogously as for the fair clients, this assumption will be relaxed in Section \ref{noise_fair_clients}.

\subsubsection{Attacks based on fixed additive stochastic perturbations}\label{sec:attacks_disguised_convergence}

%
In this new setting, we can rewrite the FedAvg aggregation process (\ref{FedAvg_server_aggregation}) for an attack with a single free-rider with perturbation $\varphi$:
\begin{equation}
\tilde{\theta}^{t+1}
=\sum_{j\in J}\frac{M_j}{N}\tilde{\theta}_j^{t+1}+\frac{M_K}{N}\tilde{\theta}^t+\frac{M_K}{N} \varphi\epsilon_{t}.
\label{eq:aggregation_disguised_fr}
\end{equation}

Theorem \ref{theo:noise_additive} extends the  results previously obtained for federated learning with plain free-riders to our new case with additive perturbations.

\begin{theorem}[Single disguised free-rider]\label{theo:noise_additive}
    Analogously to Theorem \ref{deterministic_theorem}, the aggregation process under free-riding described in equation (\ref{eq:aggregation_disguised_fr}) converges in expectation to the aggregated model of the fair clients of equation (\ref{FedAvg_server_aggregation}) :
	\begin{align}
	\E{\tilde{\theta}^t - \theta^t}
	&\xrightarrow{t\rightarrow +\infty}  0,\\
	\VAR{\tilde{\theta^t} - \theta^t}
	&\xrightarrow{t \rightarrow +\infty} \frac{ [\frac{1}{N^2} + \frac{1}{(N-M_K)^2}] \sum_{j\in J} \left( M_j\rho_j\right)^2 }{1- \left( \epsilon + \frac{M_K}{N} \right)^{2}} \nonumber \\
	& + \frac{1  }{1 - \left( \epsilon + \frac{M_K}{N} \right)^{ 2 } } \frac{M_K^2}{N^2}\varphi^2  .
    \end{align}
\end{theorem}

Theorem \ref{theo:noise_additive} shows that disguised free-riding converges to the final model of federated learning with fair clients, although with a higher variance resulting from the free-rider's perturbations injected at every iteration. The perturbation is proportional to $\frac{M_K}{N}$, the relative number of samples declared by the free-rider. 

The extension of this result to the case of multiple free-riders requires to account in equation (\ref{eq:aggregation_disguised_fr}) for an attack of the form $\sum_{k\in K} \frac{M_k}{N}\varphi_k \epsilon_{k,t}$, where $M_{k}$ is the total sample size declared by free-rider $k$. Corollary \ref{cor:noise_additive_multi} follows from the linearity of this form.


\begin{corollary}[Multiple disguised free-riders]\label{cor:noise_additive_multi}
Assuming a constant perturbation factor $\varphi_k$ for each free-rider $k$, the asymptotic expectation of Theorem \ref{theo:noise_additive} still holds, while the variance reduces to
\begin{align}
\VAR{\tilde{\theta}^t - \theta^t} 
&\xrightarrow{ t \rightarrow +\infty}    \frac{ [\frac{1}{N^2} + \frac{1}{(N-M_K)^2}] \sum_{j\in J} \left( M_j\rho_j\right)^2 }{1- \left( \epsilon + \frac{M_K}{N} \right)^{2}} \nonumber \\
& + \frac{1  }{1 - \left( \epsilon + \frac{M_K}{N} \right)^{ 2 } } \sum_{ k \in K } \frac{M_k^2}{N^2}\varphi_k^2  .
\end{align}
\end{corollary}

\subsubsection{Time-varying noise model of fair-clients evolution}\label{noise_fair_clients}	
	
To investigate more plausible  parameters evolution in federated learning, in this section we relax the assumption made in Section 
\ref{sec:sgd_perturbation} about the constant noise perturbation of the SGD process across iteration rounds. 
	

We assume here that the standard deviation ${\sigma_j^t}$ of SGD decreases at each server iteration $t$, approaching to zero over iteration rounds: ${\sigma_j^t} \xrightarrow{ t \rightarrow + \infty} 0$.  This assumption reflects the improvement of the fit of the global model $\tilde{\theta}^t$ to the local datasets over server iterations, 
and implies that the stochastic process of the local optimization of Section \ref{sec:sgd_perturbation} has noise parameter $\rho_j^t \xrightarrow{ t \rightarrow + \infty} 0$. We thus hypothesize that, to mimic the behavior of the fair clients, a suitable time-varying perturbation of the free-riders should follow a similar asymptotic behavior: $\varphi_k(t) \xrightarrow{t  \rightarrow + \infty} 0$. Under these assumptions, Corollary  \ref{cor:fair_decay_noise} shows that  the asymptotic variance of model aggregation under free-rider attacks is zero, and that it is thus still possible to retrieve the fair client's model.

\begin{corollary}\label{cor:fair_decay_noise}
Assuming that fair clients and free-riders evolve according to Section \ref{sec:sgd_perturbation} to \ref{disguised_free-rider},
if the conditions $\rho_j^t \xrightarrow{ t \rightarrow + \infty} 0$ and $\varphi_k(t) \xrightarrow{ t \rightarrow + \infty} 0 $ are met, the aggregation process of federated learning is such that the asymptotic variance of Theorems \ref{deterministic_theorem} and \ref{theo:noise_additive} reduce to 
\begin{align}
\VAR{\tilde{\theta}^t - \theta^t} \xrightarrow{ t \rightarrow + \infty} 0.
\end{align}
\end{corollary}	


We assumed in Corollary \ref{cor:fair_decay_noise} that the SGD noise ${\sigma_j^t}$ decreases at each server iteration and eventually converges to 0. In practice, the global model may not fit perfectly the dataset of the different clients $\Dcal_j$ and, after a sufficient number of optimization rounds, may keep oscillating around a local minima. We could therefore assume that ${\sigma_j^t} \xrightarrow{ t \rightarrow + \infty} \sigma_j$ leading to $\rho_j^t \xrightarrow{ t \rightarrow + \infty} \rho_j$. In this case, to mimic the behavior of the fair clients, a suitable time-varying perturbation compatible with the free-rider attacks should converge to a fixed noise level such that $\varphi_k(t) \xrightarrow{t \rightarrow + \infty} \varphi_k$. Similarly as for Corollary \ref{cor:fair_decay_noise}, it can be shown that under these hypothesis federated learning  follows the asymptotic behaviors of Theorem \ref{deterministic_theorem} and \ref{theo:noise_additive} for respectively plain and disguised free-riders. 


\subsection{FedProx}

FedProx includes a regularization term for the local loss functions of the different clients ensuring the proximity between the updated models $\theta_j^{t+1}$ and $\theta^t$. This regularization is usually defined as an additional L2 penalty term, and leads to the following form for the local gradient $g_j(\theta_j)\simeq r_j [\theta_j -\theta_j^*]+ \mu [ \theta_j - \theta^t]$ where $\mu$ is a trade-off parameter. Since the considerations in Section \ref{sec:sgd_perturbation} still hold in this setting, we can express the local model contribution for FedProx with a formulation analogous to the one of equation (\ref{eq:OU_close_form}). Hence, for FedProx, we obtain similar conclusions for Theorem \ref{deterministic_theorem} and \ref{theo:noise_additive}, as well as for Corollary \ref{cor:noise_additive_multi} and \ref{cor:fair_decay_noise}, proving that the convergence behavior with free-riders is equivalent to the one obtained with fair clients only, although with a different asymptotic variance (Appendix \ref{app:FedProx}).

\begin{theorem} Assuming convergence in expectation for federated learning with fair clients only, under the assumptions of Theorem \ref{theo:diff_plain_fr} the  asymptotic properties of  plain and disguised free-riding of  Theorem \ref{deterministic_theorem}, \ref{theo:noise_additive}, and Corollary \ref{cor:noise_additive_multi}, \ref{cor:fair_decay_noise}, still hold with FedProx. In this case we have parameters:  
\begin{equation}
{\rho_j}^2 = \frac{\lambda}{S}{\sigma_j}^2\frac{1}{2(r_j+\mu)}\left[1-e^{-2\lambda (r_j+ \mu )\frac{EM_j}{S}}\right],
\end{equation}
\begin{equation}
\epsilon = \sum_{j\in J} \frac{ M_j}{ N } [\gamma_j + \mu\frac{1 - \gamma_j}{r_j + \mu}],
\end{equation}
\begin{equation}
\text{and }\gamma_j = e^{- \lambda (r_j + \mu) \frac{E M_j}{S} }.
\end{equation}
\end{theorem}
We note that the asymptotic variance is still strictly increasing with the total number of free-riders samples. Moreover, the regularization term monitors the asymptotic variance: a higher regularization leads to a smaller noise parameter $\rho_j^2$ and to a smaller $\epsilon$, thus decreasing the asymptotic variances of Theorem \ref{deterministic_theorem}, \ref{theo:noise_additive}, and  Corollary \ref{cor:noise_additive_multi}, \ref{cor:fair_decay_noise}.

%% file: tex/Experiments.tex
\section{Experiments}
\label{experiments}
\pdfoutput=1

\begin{figure}[t]
	\centering
	\includegraphics[width=\linewidth]{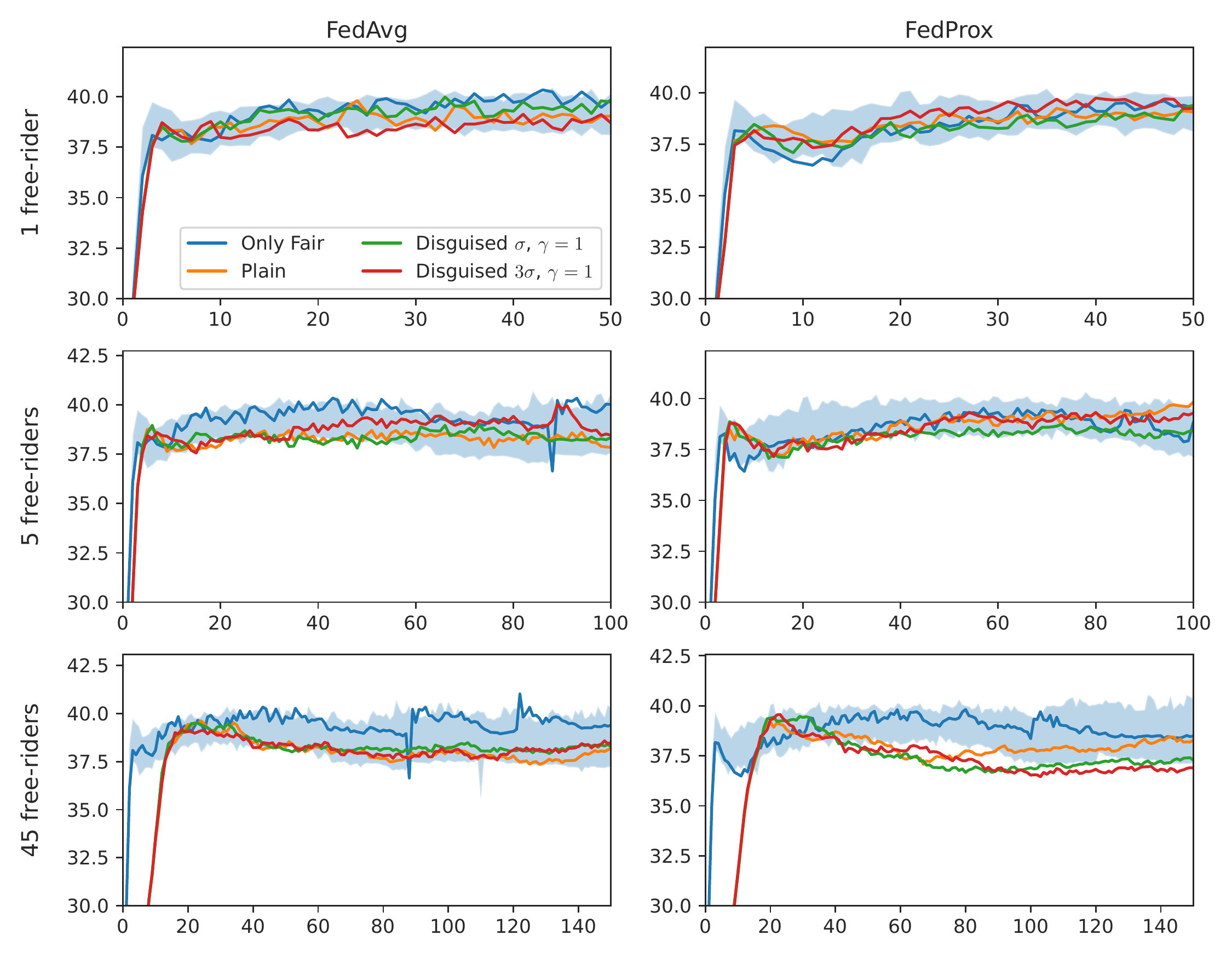}
	\caption{Plots for Shakespeare and $E=20$. Accuracy performances for FedAvg and FedProx according to the number of free-riders participating in the learning process:  15\% (top), 50\% (middle), and 90\% (bottom) of the total amount of clients. The shaded blue region indicates the variability of federated learning model with fair clients only, estimated from 30 different training initialization.}
	\label{fig:fig_1}
\end{figure}

\begin{figure}[t]
	\centering
	\includegraphics[width=\linewidth]{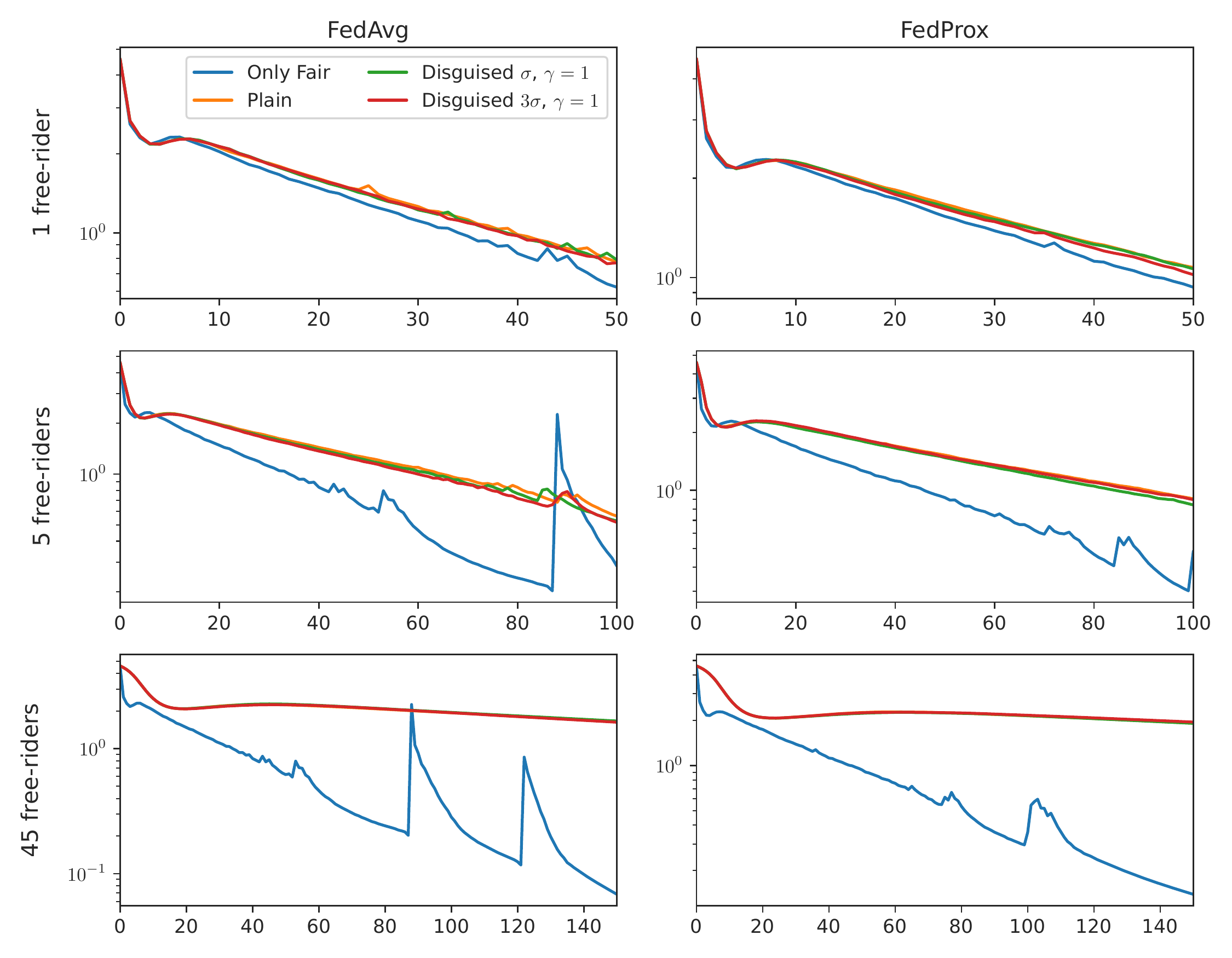}
	\caption{Plots for Shakespeare and $E=20$. Loss performances for FedAvg and FedProx according to the number of free-riders participating in the learning process:  15\% (top), 50\% (middle), and 90\% (bottom) of the total amount of clients. }
	\label{fig:fig_2}
\end{figure}

This experimental section focuses on a series of benchmarks for the proposed free-rider attacks. The methods being of general application, the focus here is to empirically demonstrate our theory on diverse experimental setups and model specifications. All code, data and experiments are available at \url{https://github.com/Accenture/Labs-Federated-Learning/tree/free-rider_attacks}. 

\subsection{Experimental Details}\label{exp:details} 

We consider 5 fair clients for each of the following scenarios, investigated in previous works on federated learning \cite{BrendanMcMahan2017,LiFedProx2018}:

\textbf{MNIST} (classification in iid and non-iid settings). We study a standard classification problem on MNIST \cite{Lecun1998} and create two benchmarks: an iid dataset (MNIST iid) where we assign 600 training digits and 300 testing digits to each client, and a non-iid dataset (MNIST non-iid), where for each digit we create two shards with 150 training samples and 75 testing samples, and allocate 4 shards for each client. For each scenario, we use a logistic regression predictor.

\textbf{CIFAR-10}\cite{CIFAR-10} (image classification). The dataset consists of 10 classes of 32x32 images with three RGB channels.  There are 50000 training examples and 10000 testing examples which we partitioned into 5 clients each containing 10000 training and 2000 testing samples.  The model architecture was taken from \cite{BrendanMcMahan2017} which consists of two convolutional layers and a linear transformation layer to produce logits.

\textbf{Shakespeare} (LSTM prediction). We study a LSTM model for next character prediction on the dataset of \textit{The Complete Works of William Shakespeare} \cite{BrendanMcMahan2017}. We randomly chose 5 clients with more than 3000 samples, and assign 70\% of the dataset to training and 30\% to testing. Each client has on average $6415.4$ samples ($\pm 1835.6$) . We use a two-layer LSTM classifier containing 100 hidden units with an 8 dimensional embedding layer. The model takes as an input a sequence of 80 characters, embeds each of the characters into a learned 8-dimensional space and outputs one character per training sample after 2 LSTM layers and a fully connected one.

We train federated models following FedAvg and FedProx aggregation processes.  In FedProx, the hyperparameter $\mu$ monitoring the regularization is chosen according to the best performing scenario reported in \cite{LiFedProx2018}: $\mu=1$ for MNIST (iid and non-iid), and $\mu=0.001$ for Shakespeare. For the free-rider we declare a number of samples equal to the average sample size across fair clients. We test federated learning with 5 and 20 local epochs using SGD optimization with learning rate $\lambda=0.001$ for MNIST (iid and non-iid), $\lambda=0.001$ for CIFAR-10, and $\lambda=0.5$ for Shakespeare, and batch size of 100.
{We evaluate the success of the free-rider attacks by quantifying testing accuracy and training loss of the resulting model, as indicators of the effect of the perturbation induced by free-riders on the final model performances.} Resulting figures for associated accuracy and loss can be found in Figure \ref{fig:fig_1}, Figure \ref{fig:fig_2} and Appendix \ref{app:plots}.

\subsection{Free-rider attacks: convergence and performances}
\label{sec:convergence_comparison}

In the following experiments, {we assume that free-riders do not have any data, which means that they cannot estimate the noise level by computing a plausible quadratic approximation of the local loss function (Section \ref{disguised_free-rider}). Therefore, }
we investigate free-rider attacks taking the simple form $\varphi(t) =  \sigma t^{-\gamma}$. The parameter $\gamma$ is chosen among a panel of testing parameters $\gamma \in \{0.5,1,2\}$, while additional experimental material on the influence of $\gamma$ on the convergence is presented in Appendix \ref{app:plots}. While the optimal tuning of disguised free-rider attacks is out of the scope of this study, in what follows the perturbations parameter $\sigma$ is defined according to practical hypotheses on the parameters evolution during federated learning.  After random initialization at the initial federated learning step, the parameter $\sigma$ is opportunely estimated to mimic the extent of the distribution of the update $\Delta \tilde{\theta}^0 = \tilde{\theta}^1 - \tilde{\theta}^0$  observed between consecutive rounds of federated learning. We can simply model these increments as a zero-centered univariate Gaussian distribution, and  assign the parameter $\sigma$ to the value of the fitted standard deviation.  According to this strategy,  the free-rider would return parameters $\tilde{\theta}_k^t$ with perturbations distributed as the ones observed between two consecutive optimization rounds. 
Figure \ref{fig:fig_1}, top row, exemplifies the evolution of the models obtained with FedAvg (20 local training epochs) on the Shakespeare dataset with respect to different scenarios: 1) fair clients only, 2) plain free-rider, 3) disguised free-rider with  decay parameter $\gamma = 1$, and estimated noise level $\sigma$, and 4) disguised free-rider with noise level increased to  $3\sigma$. For each scenario, we compare the federated model obtained under free-rider attacks with respect to the equivalent model obtained with the participation of the fair clients only. For this latter setting, to assess the model training variability, we repeated the training 30 times with different parameter  initializations.
The results show that, independently from the chosen  free-riding strategy, the resulting models attains comparable performances with respect to the one of the model obtained with fair clients only (Figure \ref{fig:fig_1}, top row). Similar results are obtained for the setup with 5 local training epochs and different values of $\gamma$, as well as for FedProx with 5 and 20 local epochs (Appendix \ref{app:plots}). 

We also investigate the same training setup 
under the influence of multiple free-riders (Figure \ref{fig:fig_1}, mid and bottom rows). 
In particular, we test the scenarios where the free-riders declare respectively  $50\%$ and $90\%$ of the total training sample size. In practice, we maintain the same experimental setting  composed of 5 fair clients, and we increase the number of  free-riders to respectively 5 and 45, while declaring for each free-rider a sample size equal to the average number of samples of the fair clients.
Independently from the magnitude of the perturbation function, the number of free-riders does not seem to affect the performance of the final aggregated model. However, the convergence speed is greatly decreased. Figure \ref{fig:fig_2} 
shows that the convergence in these different settings is not identically affected by the free-riders. When the size of free-riders is moderate, e.g. up to 50\% of the total sample size, the convergence speed of the loss is slightly slower than for federated learning with fair clients. The attacks can be still considered successful, as convergence is achieved within the pre-defined iteration  budget.  However, when the size of free-riders reaches 90\%,  convergence to the optimum is extremely slow and cannot be achieved anymore in a reasonable amount of iterations.
This result is in agreement with our theory, 
for which the convergence speed inversely proportional to the relative size of the free-riders. Interestingly, we note that the final accuracy obtained in all the scenarios is similar (though a bit slower with 90\% of free-riders), and falls within the variability observed in federated learning  with fair-clients only (Figure 1). This result is achieved in spite of the incomplete convergence during training. This effect can be explained by observing that this accuracy level is already reached at the early training stages of federated learning with fair clients, while further training does not seem to improve the predictions. This result suggests that, in spite of the very low convergence speed, the averaging process with 90\% of free-riders still achieves a reasonable minima compatible with the training path of the fair clients aggregation.

{We note that the "peaks" observed in the loss of Figure 2 are common in FL, especially in the considered application when the number of clients is low. It is important to notice that our experiments are performed by using vanilla SGD. As such, the peaks for only fair clients are to be expected in both loss and performances. We also notice that the peaks are smaller for free-riding because of the “regularization” effect of free-riders, which regresses the update towards the global model of the previous iteration.}

Analogous results and considerations can be derived from the set of experiments on the remaining datasets, training parameters and FedProx as an aggregation scheme (Appendix \ref{app:plots}).  

%% file: tex/Conclusion.tex
\section{Conclusion and discussion}
\label{conclusion}

We introduced a theoretical framework for the study of free-riding attacks on model aggregation in federated learning. Based on the proposed methodology, we proved that simple strategies based on returning the global model at each iteration already lead to successful free-rider attacks (plain free-riding), and we investigated  more sophisticated disguising techniques relying on stochastic perturbations of the parameters (disguised free-riding). The convergence of each attack was demonstrated through theoretical developments and experimental results.
{The threat of free-rider attacks is still under-investigated in machine learning. For example, current defence schemes in federated learning are mainly based on outliers detection mechanisms, to detect malicious attackers providing abnormal updates. These schemes would be therefore unsuccessful in detecting a free-rider update which is, by design, equivalent to the global federated model. 
}

This work opens the way to the investigation of optimal disguising and defense strategies for free-rider attacks, beyond the proposed heuristics. 
Our experiments show that inspection of the client's distribution should be established as a routine practice for the detection of free-rider attacks in federated learning. Further research directions are represented by the improvement of detection at the server level, through better modeling of the heterogeneity of the incoming clients' parameters. This study provides also the theoretical basis for the study of effective free-riding strategies, based on different noise model distributions and perturbation schemes. 
Finally, in this work we relied on a number of hypothesis concerning the evolution of the clients' parameters during federated learning. This choice provides us with a convenient theoretical setup for the formalization of the proposed theory which may be modified in the future, for example, for investigating more complex forms of variability and schemes for parameters aggregation.




%% file: tex/Acknowledgements.tex
\subsection*{Acknowledgments and Disclosure of Funding}
\label{sec:ack}

This work has been supported by the French government, through the 3IA Côte d’Azur Investments in the Future project managed by the National Research Agency (ANR) with the reference number ANR-19-P3IA-0002, and by the ANR JCJC project Fed-BioMed 19-CE45-0006-01. The project was also supported by Accenture.
The authors are grateful to the OPAL infrastructure from Université Côte d'Azur for providing resources and support.

%% file: tex/Supplementary_Material_alternate.tex
\newpage

\appendix

\section{Complete Proofs for FedAvg}

\subsection{Proof of Theorem \ref{theo:diff_plain_fr}}
\label{proof:diff_plain_fr}

We prove with a reasoning by induction that:
\begin{align}
\tilde{\theta}^t - \theta^t
& = \sum_{i=0}^{t-1}\left( \epsilon + \frac{M_K}{N} \right)^{t - i -1} f(\theta^i) \nonumber\\
&+ \sum_{i=0}^{t-1}\left( \epsilon + \frac{M_K}{N} \right)^{t - i -1} ( \tilde{\nu}_i - \nu_i),
\end{align}
with $f(\theta^t) 
= \frac{M_K}{N} \left[\theta^t - \sum_{j\in J} \frac{M_j}{N - M_K} [\eta_j (\theta^t -\theta_j^*) + \theta_j^*]\right]$, 
$\epsilon = \sum_{j\in J} \frac{ M_j}{ N } \eta_j$, $\nu_t = \sum_{j\in J} \frac{ M_j}{ N - M_K} \rho_j \zeta_{j, t}$
and $ \tilde{\nu}_t= \sum_{j\in J} \frac{ M_j}{ N } \rho_j \tilde{\zeta}_{j, t}$. By definition of $\theta^{t+1}$, $\E{f(\theta^t )} 
	= \frac{M_K}{N} \left[\E{\theta^t} - \E{\theta^{t+1}} \right] $.

\begin{proof}
\underline{\textbf{Server iteration $t=1$}}

Using the fair clients local model parameters evolution of Section \ref{sec:sgd_perturbation} and the server aggregation process expressed in equation (\ref{neutral_server_aggregation}), the global model can be written as
\begin{align}
\label{app:eq:ini_FL}
\theta^1
= \sum_{j\in J}\frac{M_j}{N-M_K}\left[ \eta_j \left( \theta^0 - \theta_j^* \right) + \theta_j^* \right]
+ \nu_0.
\end{align}
Similarly, the global model for federated learning with plain free-riders can be expressed as
\begin{align}
\label{app:eq:ini_FR}
\tilde{\theta}^1
= \sum_{j\in J}\frac{M_j}{N}\left[ \eta_j \left( \theta^0 - \theta_j^* \right) + \theta_j^* \right]
+\frac{M_K}{N}\theta^0
+ \tilde{\nu}_0.
\end{align}
By subtracting equation (\ref{app:eq:ini_FL}) to equation (\ref{app:eq:ini_FR}), we obtain:
\begin{align}
\tilde{\theta}^1 - \theta^1
&= -\frac{M_K}{N }\sum_{j\in J}\frac{M_j}{N - M_K}\left[ \eta_j \left( \theta^0 - \theta_j^* \right) + \theta_j^* \right] \nonumber\\
&+\frac{M_K}{N}\theta^0
+ \tilde{\nu}_0 - \nu_0
\end{align}
Hence, $\tilde{\theta}_1 - \theta_1$ follows the formalization.

\underline{\textbf{From $t$ to $t+1$}}

We suppose the property true at a server iteration $t$. Hence, we get:
\begin{align}
\label{app:eq:assumption}
\tilde{\theta}^t - \theta^t
& = \sum_{i=0}^{t-1}\left( \epsilon + \frac{M_K}{N} \right)^{t - i -1} f(\theta^i) \nonumber\\
&+ \sum_{i=0}^{t-1}\left( \epsilon + \frac{M_K}{N} \right)^{t - i -1} ( \tilde{\nu}_i - \nu_i),
\end{align}

With the same reasoning as for $t=1$, we get:
\begin{align}
\label{app:eq:induc_FL_n}
\theta^{t+1}
= \sum_{j\in J}\frac{M_j}{N-M_K}\left[ \eta_j \left( \theta^t - \theta_j^* \right) + \theta_j^* \right]
+ \nu_t
\end{align}
and 
\begin{align}
\label{app:eq:induc_FR_n}
\tilde{\theta}^{t+1}
= \sum_{j\in J}\frac{M_j}{N}\left[ \eta_j \left( \tilde{\theta}^t - \theta_j^* \right) + \theta_j^* \right]
+\frac{M_K}{N}\tilde{\theta}^t
+ \tilde{\nu}_t
\end{align}
By using equation (\ref{app:eq:assumption}) for equation (\ref{app:eq:induc_FR_n}), we get:
\begin{align}
\tilde{\theta}^{t+1}
&= \sum_{j\in J}\frac{M_j}{N}\left[ \eta_j \left( {\theta}^t - \theta_j^* \right) + \theta_j^* \right] \nonumber\\
& + \epsilon \sum_{i=0}^{t-1}\left( \epsilon + \frac{M_K}{N} \right)^{t - i -1} f(\theta^i) \nonumber\\
&+ \epsilon \sum_{i=0}^{t-1}\left( \epsilon + \frac{M_K}{N} \right)^{t - i -1} ( \tilde{\nu}_i - \nu_i) \nonumber\\
&+\frac{M_K}{N}{\theta}^t \nonumber\\
& + \frac{M_K}{N}\sum_{i=0}^{t-1}\left( \epsilon + \frac{M_K}{N} \right)^{t - i -1} f(\theta^i) \nonumber\\
&+ \frac{M_K}{N}\sum_{i=0}^{t-1}\left( \epsilon + \frac{M_K}{N} \right)^{t - i -1} ( \tilde{\nu}_i - \nu_i) \nonumber\\
&+ \tilde{\nu}_t 
\end{align}
which can be rewritten as:
\begin{align}
\tilde{\theta}^{t+1}
&= \sum_{j\in J}\frac{M_j}{N}\left[ \eta_j \left( {\theta}^t - \theta_j^* \right) + \theta_j^* \right] \nonumber\\
& + [\epsilon + \frac{M_K}{N}] \sum_{i=0}^{t-1}\left( \epsilon + \frac{M_K}{N} \right)^{t - i -1} f(\theta^i) \nonumber\\
&+  [\epsilon + \frac{M_K}{N}] \sum_{i=0}^{t-1}\left( \epsilon + \frac{M_K}{N} \right)^{t - i -1} ( \tilde{\nu}_i - \nu_i) \nonumber\\
&+\frac{M_K}{N}{\theta}^t + \tilde{\nu}_t ,
\end{align}
leading to 
\begin{align}
\label{app:eq:induc_FR_n_changed}
\tilde{\theta}^{t+1}
&= \sum_{j\in J}\frac{M_j}{N}\left[ \eta_j \left( {\theta}^t - \theta_j^* \right) + \theta_j^* \right] \nonumber\\
& + \sum_{i=0}^{t-1}\left( \epsilon + \frac{M_K}{N} \right)^{t-i} f(\theta^i) \nonumber\\
&+   \sum_{i=0}^{t-1}\left( \epsilon + \frac{M_K}{N} \right)^{t-i} ( \tilde{\nu}_i - \nu_i) \nonumber\\
&+\frac{M_K}{N}{\theta}^t + \tilde{\nu}_t 
\end{align}
By subtracting equation (\ref{app:eq:induc_FR_n_changed}) to equation (\ref{app:eq:induc_FL_n}), we obtain:
\begin{align}
\tilde{\theta}^{t+1} - \theta^{t+1}
&= -\frac{M_K}{N }\sum_{j\in J}\frac{M_j}{N - M_K}\left[ \eta_j \left( \theta^t - \theta_j^* \right) + \theta_j^* \right] \nonumber\\
& + \sum_{i=0}^{t-1}\left( \epsilon + \frac{M_K}{N} \right)^{t-i} f(\theta^i) \nonumber\\
&+   \sum_{i=0}^{t-1}\left( \epsilon + \frac{M_K}{N} \right)^{t-i} ( \tilde{\nu}_i - \nu_i) \nonumber\\
&+\frac{M_K}{N}{\theta}^t + \tilde{\nu}_t - \nu_t
\end{align}
Given that $-\frac{M_K}{N }\sum_{j\in J}\frac{M_j}{N - M_K}\left[ \eta_j \left( \theta^t - \theta_j^* \right) + \theta_j^* \right] +\frac{M_K}{N} \theta^t=f(\theta^t)$, we get:
\begin{align}
\tilde{\theta}^{t+1} - \theta^{t+1}
&= \sum_{i=0}^{t}\left( \epsilon + \frac{M_K}{N} \right)^{t-i} f(\theta^i) \nonumber\\
&+   \sum_{i=0}^{t}\left( \epsilon + \frac{M_K}{N} \right)^{t-i} ( \tilde{\nu}_i - \nu_i) .
\end{align}
\end{proof}

\subsection{Proof of Theorem \ref{deterministic_theorem}}
\label{deterministic_proof}

\begin{proof}
\underline{\textbf{Expected Value}}

Let us first have a look at the expected value. By definition, a sum of Gaussian distributions with 0 mean, $\E{ \nu_i} = 0$ and $\E{ \tilde{\nu}_i} = 0$. We also notice that $\E{f(\theta^t )} 
	= \frac{M_K}{N} \left[\E{\theta^t} - \E{\theta^{t+1}} \right] $. Hence, we obtain
\begin{align}
\label{app:eq:expected_value_original_expression}
\E{\tilde{\theta}^t - \theta^t}
& = \frac{M_K}{N} \sum_{i=0}^{t-1}\left( \epsilon + \frac{M_K}{N} \right)^{n- i -1} \E{\theta^t - \theta^{t+1}} .
\end{align}
We consider that federated learning is converging, hence $|\E{\theta^t} - \E{\theta^{t+1}}| \xrightarrow{t  \rightarrow + \infty}0$, and for any positive $\alpha$, there exists $N_0$ such that $|\E{\theta^t - \theta^{t+1}}| < \alpha $. Since $\eta_j \in ]0,1[$, we have  $\epsilon \in ]0, \frac{N-M_K}{N}[$ and  $\epsilon + \frac{M_K}{N} \in ]0, 1[$. Thus, we can rewrite equation (\ref{app:eq:expected_value_original_expression}) as
\begin{align}
|\E{\tilde{\theta}^t - \theta^t}|
& \le  \sum_{i=0}^{N_0-1}\left( \epsilon + \frac{M_K}{N} \right)^{t - i -1} | \E{\theta^t} - \E{\theta^{t+1}} | \nonumber\\
& + \sum_{i=N_0}^{t-1}\left( \epsilon + \frac{M_K}{N} \right)^{t - i -1} \alpha .
\end{align}
We define by $R_\alpha=\max_{i\in [1,N_0]} | \E{\theta^t} - \E{\theta^{t+1}} |$, and get:
\begin{align}
|\E{\tilde{\theta}^t - \theta^t}|
& \le \underbrace{ \sum_{i=0}^{N_0-1}\left( \epsilon + \frac{M_K}{N} \right)^{t - i -1}}_A R_\alpha \nonumber\\
& + \underbrace{\sum_{i=N_0}^{t-1}\left( \epsilon + \frac{M_K}{N} \right)^{t - i -1}}_B \alpha . \label{app:eq:E_diff}
\end{align}

\begin{itemize}
    
    \item \textbf{Expressing $A$.}
    \begin{align}
    A
    &= \sum_{i=0}^{N_0-1}\left( \epsilon + \frac{M_K}{N} \right)^{t - i -1} \\
    & = \left( \epsilon + \frac{M_K}{N} \right)^{t-1}\frac{1 - \left( \epsilon + \frac{M_K}{N} \right)^{-N_0}}{1 -\left( \epsilon + \frac{M_K}{N} \right)^{-1} } \\
    & \xrightarrow{ t  \rightarrow + \infty} 0
    \label{app:eq:A_final}
    \end{align}

    \item \textbf{Expressing $B$.}
    \begin{align}
    B
    &=\sum_{i=N_0}^{t-1}\left( \epsilon + \frac{M_K}{N} \right)^{t - i -1}\\
    & = \left( \epsilon + \frac{M_K}{N} \right)^{t - N_0 - 1}\frac{1 - \left( \epsilon + \frac{M_K}{N} \right)^{-(t-N_0)}}{1 - \left( \epsilon + \frac{M_K}{N} \right)^{-1} } \\
    & = \frac{1 - \left( \epsilon + \frac{M_K}{N} \right)^{t-N_0}}{1 - \left( \epsilon + \frac{M_K}{N} \right) } \\
    & \xrightarrow{t \rightarrow + \infty} \frac{1}{1 - \left( \epsilon + \frac{M_K}{N} \right) } > 0 \label{app:eq:B_final}
    \end{align}

\end{itemize}

Using equation (\ref{app:eq:A_final}) and (\ref{app:eq:B_final}) in equation (\ref{app:eq:E_diff}), we get:
\begin{align}
\forall  \alpha \lim_{t  \rightarrow + \infty } |\E{\tilde{\theta}^t - \theta^t}| \le B \alpha ,
\end{align}
which is equivalent to
\begin{align}
\lim_{t  \rightarrow + \infty } \E{\tilde{\theta}^t - \theta^t} = 0 .
\end{align}

\underline{\textbf{Variance}}

The Wiener processes, $\nu_i$ and $\tilde{\nu}_i$ are independent from the server models parameters $\theta^i$. Also, each Wiener process is independent with the other Wiener processes. Hence, we get:
\begin{align}
\label{app:eq:variance_plain}
\VAR{\tilde{\theta}^t - \theta^t}
& = \underbrace{\VAR{\sum_{i=0}^{t-1}\left( \epsilon + \frac{M_K}{N} \right)^{t - i -1} f(\theta^i)}}_E \nonumber\\
&+ \sum_{i=0}^{t-1}\left( \epsilon + \frac{M_K}{N} \right)^{2(t - i -1) } \underbrace{\VAR{\tilde{\nu}_i - \nu_i}}_F,
\end{align}

\textbf{Expressing $E$.} Before getting a simpler expression for $E$, we need to consider $\COV{f(\theta^l),f(\theta^m)}$. To do so, we first consider $f( \theta^t)-\E{f(\theta^t )}$.
\begin{align}
&f(\theta^t )-\E{f(\theta^t)} \nonumber\\
&= \underbrace{\frac{M_K}{N}\left[ 1 - \sum_{j\in J} \frac{M_j}{ N - M_K } \eta_j \right]}_G [\theta^t - \E{\theta^t}] , 
\end{align}
We can prove with a reasoning by induction that $\theta^t-\E{\theta^t}=\sum_{i=0}^{n - 1} \left(\sum_{j\in J} \frac{M_j}{N-M_K} \eta_j\right)^{t - i -1} \nu_i = \sum_{k=0}^{n - 1} \epsilon^{t - i -1} \nu_i$. All the $\nu_i$ are independent across each others and have 0 mean, hence:
\begin{align}
\COV{f(\theta_l ),f(\theta_m)} 
= G^2 \sum_{i=0}^{\min \{l-1,m-1\} } \epsilon^{l + m - 2i - 2 } \E{  \nu_i^2} 
\end{align}
Considering that $\E{  \nu_i^2}= \VAR{\nu_i} = \sum_{j\in J} \left( \frac{M_j}{N - M_K} \rho_j \right)^2$, we get:
\begin{align}
&\COV{f(\theta^l ),f(\theta^m )} \nonumber\\
&= G^2 \sum_{j\in J} \left( \frac{M_j}{N - M_K} \rho_j \right)^2 \sum_{i=0}^{\min \{ l-1, m-1\}} \epsilon^{t - i -1}
\end{align}
We define $G'= G^2\sum_{j\in J}\left( \frac{M_j}{N - M_K} \rho_j \right)^2$. Given that $\epsilon\in ]0,1[$, we get the following upper bound on $E$:
\begin{align}
\COV{f(\theta^l),f(\theta^m)} \le G' \min \{ l, m\}
\end{align}
By denoting $H= \epsilon + \frac{M_K}{N} $, we can rewrite $E$ as:
\begin{align}
E
& = \sum_{l=0}^{t-1} \sum_{m=0}^{t-1} H^{2(t-1)- l-m } \COV{f_l(\theta^l), f(\theta^m)}\\
& \le \sum_{l=0}^{t-1} \sum_{m=0}^{t-1} H^{2(t-1)-l-m } G' \min \{ l, m \} 
\end{align}
Considering that $\min \{ l, m \}\le l $, we get:
\begin{align}
E
& \le G' \sum_{l=0}^{t-1} \sum_{m=0}^{t-1} H^{2 (t-1) -l - m } l\\
&=   G' H^{2 (t-1)}\sum_{l=0}^{t-1} H^{ -l } l \sum_{m=0}^{t-1} H^{ - m } \\
&=  G' H^{2 (t-1)}\sum_{l=0}^{t-1} H^{-l}  l \frac{1 - H^{-n}}{1-H^{-1}}\\
&=  G' H^{2 (t-1)}\frac{1 - H^{-n}}{1-H^{-1}}\sum_{l=0}^{t-1}H^{-l } l \label{app:eq:final_E} 
\end{align} 
Considering the power series $\sum_{k=0}^{+\infty} n x^n= \frac{x}{(1-x)^2}$, we get that $\sum_{l=0}^{t-1}H^{-l } l= \frac{H^{-1}}{( 1 - H^{-1} )^2}$ . Hence, $E$'s upper bound goes to 0. Given that $E$ is non-negative, we get:
\begin{equation}
E \xrightarrow{ t  \rightarrow + \infty} 0
\end{equation}

\textbf{Expressing $F$.} Let us first consider the noise coming from the SGD steps. All the $\tilde{\nu}_i$ are independent with $\nu_i$. Hence, we have 
\begin{align}
F
&= \VAR{\tilde{\nu}_i}  - \VAR{\nu_i }\\
& = \VAR{\sum_{j\in J} \frac{ M_j}{ N } \rho_j \tilde{\zeta}_{j,i} - \sum_{j\in J} \frac{ M_j}{ N - M_K } \rho_j {\zeta}_{j,i}}  \\
&= [\frac{1}{N^2} + \frac{1}{(N-M_K)^2}]\sum_{j\in J} \left( M_j\rho_j\right)^2\label{app:eq:final_F}
\end{align}

Replacing (\ref{app:eq:final_F}) in equation (\ref{app:eq:variance_plain}), we can express the variance as
\begin{align}
\VAR{\tilde{\theta}^t - \theta^t}
& = E + F  \sum_{i=0}^{t-1} H^{2( t - i -1)}  \\
& = E + F  H^{2( t - 1)}\sum_{i=0}^{t-1} H^{ -2  i }  \\
& = E + F  H^{2( t - 1)}  \frac{ 1 - H^{-2 t}}{1- H^{-2}} \\
& = E + F  \frac{ 1- H^{2 t}   }{1- H^{2}} 
\end{align}
By replacing $F$ and $H$ with their respective expression, we can conclude that 
\begin{equation}
\VAR{\tilde{\theta}^t - \theta^t} 
\xrightarrow{ t  \rightarrow + \infty}    \frac{ [\frac{1}{N^2} + \frac{1}{(N-M_K)^2}] \sum_{j\in J} \left( M_j\rho_j\right)^2 }{1- \left( \epsilon + \frac{M_K}{N} \right)^{2}}
\end{equation}

\end{proof}

\underline{Note 1:} The asymptotic variance is strictly increasing with the number of data points declared by the free-riders $M_K$.

While $M_j$ and $\rho_j$ are constants and independent from the number of free-riders and from their respective number of data points, $N$ and $\epsilon$ depend on the total number of free-riders' samples $M_K$. We first rewrite $\epsilon = \frac{1}{N}\alpha $ with $\alpha = \sum_{j \in J} M_j \eta_j$ not depending on $M_K$ and we get:
\begin{equation}
\epsilon + \frac{M_K}{N}
= \frac{1}{N}[\alpha + M_K].
\end{equation}
By defining $M_J=\sum_{j \in J} M_j$, we get:
\begin{align}
1- \left( \epsilon  + \frac{M_K}{N} \right)^2
= \frac{1}{N^2} [ M_J^2 + 2 M_K[M_J - \alpha] - \alpha^2 ],
\end{align}
with $M_J - \alpha>0$ because $\eta_j \in ]0,1[$.

Also, considering that 
\begin{align}
\frac{1}{N^2} + \frac{1}{(N- M_K)^2}
= \frac{1}{N^2}[\frac{M_K^2}{M_J^2} + 2 \frac{M_K}{M_J} + 2],
\end{align}
we can rewrite 
\begin{align}
\frac{ \frac{1}{N^2} + \frac{1}{(N-M_K)^2} }{1- \left( \epsilon + \frac{M_K}{N} \right)^{2}}
= \frac{\frac{M_K^2}{M_J^2} + 2 \frac{M_K}{M_J} + 2}{ M_J^2 + 2 M_K[M_J - \alpha] - \alpha^2}
\end{align}
As the numerator is a polynomial of order 2 in $M_K$ and the denominator is a polynomial of order 1 in $M_K$, the asymptotic variance is increasing with $M_K$.

\underline{Note 2:} When considering that the SGD noise variance is different for federated learning with and without free-riders, we get:
\begin{align}
F
&= \frac{1}{N^2} \sum_{j\in J}\left( M_j\tilde{\rho}_j\right)^2+ \frac{1}{(N-M_K)^2}\sum_{j\in J} \left( M_j\rho_j\right)^2
\end{align}

\subsection{Proof of Theorem \ref{theo:noise_additive}}
\label{stochastic_proof_additive}

\begin{proof}

\textbf{Relation between federated learning with and without free-riders global model}

With a reasoning by induction similar to Proof \ref{proof:diff_plain_fr}, we get:
\begin{align}
\tilde{\theta}^t - \theta^t
& = \sum_{i=0}^{t-1}\left( \epsilon + \frac{M_K}{N} \right)^{t - i -1} f(\theta^i)\\
&+ \sum_{i=0}^{t-1}\left( \epsilon + \frac{M_K}{N} \right)^{t - i -1} ( \tilde{\nu}_i - \nu_i)\\
&+ \sum_{i=0}^{t-1}\left( \epsilon + \frac{M_K}{N} \right)^{t - i -1} \frac{M_K}{N}\varphi \epsilon_t,
\end{align}

\textbf{Expected value}

$\epsilon_t$ is a delta-correlated Gaussian White noise which implies that $\E{\epsilon_t} =0 $. Following the same reasoning steps as in Proof \ref{deterministic_proof}, we get:
\begin{align}
\lim_{t  \rightarrow + \infty } \E{\tilde{\theta}^t - \theta^t} = 0 .
\end{align}

\textbf{Variance}

All the $\epsilon_t$ are independent Gaussian white noises implying $\VAR{\epsilon_t}=1$. Following the same reasoning steps as in Proof \ref{deterministic_proof}, we get:
\begin{align}
&\VAR{ \sum_{i=0}^{t-1}\left( \epsilon + \frac{M_K}{N} \right)^{t - i -1} \frac{M_K}{N}\varphi \epsilon_t } \nonumber\\
&=   \sum_{i=0}^{t-1}\left( \epsilon + \frac{M_K}{N} \right)^{2 (t - i -1) } \frac{M_K^2}{N^2}\varphi^2  \\
&=   \left( \epsilon + \frac{M_K}{N} \right)^{2 (t  -1) } \frac{1 - \left( \epsilon + \frac{M_K}{N} \right)^{-2 t } }{1 - \left( \epsilon + \frac{M_K}{N} \right)^{ -2 } } \frac{M_K^2}{N^2}\varphi^2  \\
&=   \frac{1 - \left( \epsilon + \frac{M_K}{N} \right)^{2 t } }{1 - \left( \epsilon + \frac{M_K}{N} \right)^{ 2 } } \frac{M_K^2}{N^2}\varphi^2  \\
&\xrightarrow{ t \rightarrow + \infty}   \frac{1  }{1 - \left( \epsilon + \frac{M_K}{N} \right)^{ 2 } } \frac{M_K^2}{N^2}\varphi^2  
\end{align}
As for equation (\ref{app:eq:variance_plain}), all the $\epsilon_t$ are independent from $\nu_t$, from $\tilde{\nu}_t$, and from the global model parameters $\theta^t$. Hence, for one disguised free-rider we get the following asymptotic variance:
\begin{align}
\VAR{\tilde{\theta}^t - \theta^t} 
&\xrightarrow{ t \rightarrow +\infty}    \frac{ [\frac{1}{N^2} + \frac{1}{(N-M_K)^2}] \sum_{j\in J} \left( M_j\rho_j\right)^2 }{1- \left( \epsilon + \frac{M_K}{N} \right)^{2}} \nonumber \\
& + \frac{1  }{1 - \left( \epsilon + \frac{M_K}{N} \right)^{ 2 } } \frac{M_K^2}{N^2}\varphi^2  .
\end{align}

\end{proof}

\subsection{Proof of Corollary \ref{cor:noise_additive_multi}}
\label{proof:disguised_multi}

\begin{proof}

\textbf{Relation between federated learning with and without free-riders global model}

With a reasoning by induction similar to Proof \ref{proof:diff_plain_fr}, we get:
\begin{align}
\tilde{\theta}^t - \theta^t
& = \sum_{i=0}^{t-1}\left( \epsilon + \frac{M_K}{N} \right)^{t - i -1} f(\theta^i)\\
&+ \sum_{i=0}^{t-1}\left( \epsilon + \frac{M_K}{N} \right)^{t - i -1} ( \tilde{\nu}_i - \nu_i)\\
&+ \sum_{k \in K} \sum_{i=0}^{t-1}\left( \epsilon + \frac{M_K}{N} \right)^{t - i -1} \frac{M_k}{N}\varphi_k \epsilon_{k,t},
\end{align}

\textbf{Expected value}

$\epsilon_{k,t}$ are delta-correlated Gaussian White noises which implies that $\E{\epsilon_{k,t}} =0 $. Following the same reasoning steps as in Proof \ref{deterministic_proof}, we get:
\begin{align}
\lim_{t  \rightarrow + \infty } \E{\tilde{\theta}^t - \theta^t} = 0 .
\end{align}

\textbf{Variance}

All the $\epsilon_{k, t}$ are independent Gaussian white noises over server iterations $t$ and free-riders indices $k$ implying $\VAR{\epsilon_t}=1$. Following the same reasoning steps as in Proof \ref{deterministic_proof}, we get:
\begin{align}
\VAR{ \sum_{i=0}^{t-1}\left( \epsilon + \frac{M_K}{N} \right)^{t - i -1} \frac{M_k}{N}\varphi_k \epsilon_{k, t} } \nonumber\\
\xrightarrow{ t \rightarrow + \infty}   \frac{1  }{1 - \left( \epsilon + \frac{M_K}{N} \right)^{ 2 } } \frac{M_k^2}{N^2}\varphi_k^2  
\end{align}

Like for equation (\ref{app:eq:variance_plain}), all the $\epsilon_{k,t}$ are independent from $\nu_t$, $\tilde{\nu}_t$ and the global model parameters $\theta^t$. Hence, for multiple disguised free-rider we get the following asymptotic variance:
\begin{align}
\VAR{\tilde{\theta}^t - \theta^t} 
&\xrightarrow{ t \rightarrow +\infty}    \frac{ [\frac{1}{N^2} + \frac{1}{(N-M_K)^2}] \sum_{j\in J} \left( M_j\rho_j\right)^2 }{1- \left( \epsilon + \frac{M_K}{N} \right)^{2}} \nonumber \\
& + \frac{1  }{1 - \left( \epsilon + \frac{M_K}{N} \right)^{ 2 } } \sum_{ k \in K } \frac{M_k^2}{N^2}\varphi_k^2  .
\end{align}

\end{proof}

\subsection{Proof of Corollary \ref{cor:fair_decay_noise}}
\label{proof:fair_decay_noise}

\begin{proof}

\textbf{Relation between federated learning with and without free-riders global model}

The relation remains the same for Theorem \ref{deterministic_theorem}, Theorem \ref{theo:noise_additive}, and Corollary \ref{cor:noise_additive_multi} by replacing $\eta_j$ with $\eta_j(t) = \sum{j \in J} \frac{M_j}{N} \rho_j(t)$ and $\varphi_k$ by $\varphi_k(t)$ for disguised free-riding.

\textbf{Expected value}

With $\rho_j^t$ and $\varphi(t)$ the properties for $\tilde{\nu}_t$, $\nu_t$, $\epsilon_t$ and $\epsilon_{k,t}$ remain identical. Hence, they still are delta-correlated Gaussian White noises implying that $\E{\tilde{\nu}_t} = \E{ {\nu}_t} = \E{\epsilon_{t} } = \E{\epsilon_{k,t} }=0 $. Hence, for Theorem \ref{deterministic_theorem}, Theorem \ref{theo:noise_additive}, and Corollary \ref{cor:noise_additive_multi}, we get:
\begin{align}
\lim_{t  \rightarrow + \infty } \E{\tilde{\theta}^t - \theta^t} = 0 .
\end{align}

\textbf{Variance}

Variance asymptotic behaviour proven in Proof \ref{deterministic_proof}, \ref{stochastic_proof_additive}, and \ref{proof:disguised_multi} can be reduced to the one in Proof \ref{deterministic_proof}. Hence, $F$, equation (\ref{app:eq:final_F}), need to be reexpressed to take into account $\rho_j(t)$. All the $\tilde{\nu}_i$ are still independent with $\nu_i$. Hence, we have:
\begin{align}
F
&= \VAR{\tilde{\nu}_i(t)  -\nu_i(t) }\\
& = \VAR{\sum_{j\in J} \frac{ M_j}{ N } \rho_j^t \tilde{\zeta}_{j,i} - \sum_{j\in J} \frac{ M_j}{ N - M_K } \rho_j^t {\zeta}_{j,i}}  
\end{align}
Considering that $ \rho_j^t \xrightarrow{ t \rightarrow + \infty} 0$, we get:
\begin{align}
F \xrightarrow{ t \rightarrow + \infty}   0
\end{align}
Using the same reasoning as the one used for the expected value convergence in Proof \ref{deterministic_proof}, we get that the SGD noise contribution linked to $F$ goes to 0 at infinity.

For the disguised free-riders,  $\epsilon_{k,t}$ are still independent Gaussian white noises implying $\VAR{\epsilon_{k,t} } = 1$. Hence, following a reasoning similar to the on in Proof \ref{deterministic_proof}, we get:
\begin{align}
\VAR{ \sum_{i=0}^{t-1}\left( \epsilon + \frac{M_K}{N} \right)^{t - i -1} \frac{M_K}{N}\varphi_k(t) \epsilon_{k, t} } \nonumber\\
=   \sum_{i=0}^{t-1}\left( \epsilon + \frac{M_K}{N} \right)^{2 (t - i -1) } \frac{M_K^2}{N^2}\varphi_k^2(t)  
\end{align}
Considering that $ \varphi_k(t) \xrightarrow{ t \rightarrow + \infty} 0$, by using the same reasoning as for the proof of the expected value for free-riders, Section XX, we get:
\begin{align}
\VAR{ \sum_{i=0}^{t-1}\left( \epsilon + \frac{M_K}{N} \right)^{t - i -1} \frac{M_K}{N}\varphi_k(t) \epsilon_{k, t} } 
\xrightarrow{ t \rightarrow + \infty}   0
\end{align}
Hence, we can conclude that
\begin{align}
\VAR{\tilde{\theta}^t - \theta^t} 
&\xrightarrow{ t \rightarrow +\infty}  0.
\end{align}

\end{proof}

\section{Complete Proofs for FedProx}
\label{app:FedProx}

FedProx is a generalization of FedAvg. As such, we use the proof done for FedAvg to prove convergence of free-riders attack using FedProx as an optimization solver. The L2 norm monitored by $\mu$ changes the gradient as $g_j(\theta_j)\simeq r_j [\theta_j -\theta_j^*]+ \mu [ \theta_j - \theta^t]$.

Using equation (\ref{eq:SDE_SGD}), we then get:
\begin{align}
\mathrm{d} \theta_j = -\lambda \left[r_j [\theta_j - \theta_j^*]+ \mu [ \theta_j - \theta^t]\right] + \frac{\lambda}{\sqrt{S}} \sigma_j(\theta_j) \mathrm{d}W_j,
\end{align}
leading to
\begin{align}
\theta_j(u)
& =e^{-\lambda[r_j+\mu] u} \theta_j(0) +\frac{r_j \theta_j^* + \mu \theta^t}{r_j + \mu} [1 - e^{-\lambda (r_j + \mu) u}] \nonumber\\
& + \frac{\lambda }{\sqrt{S}} \int_{x=0}^u e^{-\lambda (r_j + \mu) ( u -x ) }\sigma_j (\theta_j) \mathrm{d}W_x .
\end{align}
considering that $\theta_j(0)=\theta^t$, $\theta_j(\frac{E M_j}{S})= \theta_j^{t+1}$, and $\sigma_j(\theta_j)=\sigma_j^t$, we get:
\begin{align}
\theta_j^{t+1}
& = \gamma_j \theta^t +\frac{r_j \theta_j^* + \mu \theta^t}{r_j + \mu} [1 - \gamma_j]\\
& + \frac{\lambda }{\sqrt{S}} \int_{x=0}^{\frac{E M_j}{S}} e^{-\lambda (r_j + \mu) (\frac{E M_j}{S} - x) }\sigma_j^t \mathrm{d}W_x,
\end{align}
where $\gamma_j= e^{-\lambda[r_j+\mu] \frac{E M_j}{S}}$.
We can reformulate this as
\begin{align}
\theta_j^{t+1}
& = [\gamma_j +\mu \frac{1- \gamma_j}{r_j + \mu}] \theta^t +\frac{r_j }{r_j + \mu} [1 - \gamma_j]\theta_j^*\\
& + \frac{\lambda }{\sqrt{S}} \int_{x=0}^{\frac{E M_j}{S}} e^{-\lambda (r_j + \mu) (\frac{E M_j}{S} - x) }\sigma_j^t \mathrm{d}W_x,
\end{align}
The SGD noise variance between two server iterations for FedProx is:
\begin{align}
\VAR{{\theta}_j^{t+1}|\theta^t}
&=\underbrace{\frac{\lambda}{S}{\sigma_j^t}^2\frac{1}{2(r_j+\mu)}\left[1-e^{-2\lambda (r_j+ \mu )\frac{EM_j}{S}}\right]}_{{\rho_j^t}^2},
\label{eq:variance_SGD_FP}
\end{align}

We also define $\eta_j' = \gamma_j +\mu \frac{1- \gamma_j}{r_j + \mu}$ and $\delta_j = \frac{r_j }{r_j + \mu} [1 - \gamma_j]$. For FedAvg, $\mu=0$, we get $\eta_j' = \eta_j$ and $\delta_j = 1 - \eta_j$. By property of the exponential, $\gamma_j \in ]0,1[$. As $r_j$ and $\mu$ are non negative, then $\eta_j'\in ]0,1[$ like $\eta_j$ for FedAvg.

\textbf{Theorem \ref{theo:diff_plain_fr} for FedProx}

We consider ${\rho_j'}^2 = \frac{\lambda}{S}{\sigma_j}^2\frac{1}{2(r_j+\mu)}\left[1-e^{-2\lambda (r_j+ \mu )\frac{EM_j}{S}}\right]$

Using the same reasoning by induction as in Proof \ref{proof:diff_plain_fr}, we get:
\begin{align}
\tilde{\theta}^t - \theta^t
& = \sum_{i=0}^{t-1}\left( \epsilon' + \frac{M_K}{N} \right)^{t - i -1} g(\theta^i) \nonumber\\
&+ \sum_{i=0}^{t-1}\left( \epsilon' + \frac{M_K}{N} \right)^{t - i -1} ( \tilde{\nu}_i' - \nu_i' ),
\end{align}
with $g(\theta^t) = \frac{M_K}{N} \left[\theta^t - \sum_{j\in J} \frac{M_j}{N - M_K} [\eta_j' \theta^t + \delta_j \theta_j^*]\right]$, 
$\epsilon' = \sum_{j\in J} \frac{ M_j}{ N } \eta_j'$, $\nu_t' = \sum_{j\in J} \frac{ M_j}{ N - M_K} \rho_j' \zeta_{j, t}$
and $ \tilde{\nu}_t' = \sum_{j\in J} \frac{ M_j}{ N } \rho_j' \tilde{\zeta}_{j, t}$.

\textbf{Theorem \ref{deterministic_theorem} for FedProx}

Like for FedAvg, we make the assumption that federated learning without free-riders using FedProx converge. In addition, $ \tilde{\nu}_t'$ and $\nu_t'$ are also independent delta-correlated Gaussian white noises. Following the same proof as in Proof \ref{deterministic_proof}, we thus get:
\begin{align}
\lim_{t  \rightarrow + \infty } \E{\tilde{\theta}^t - \theta^t} = 0 .
\end{align}
and  
\begin{equation}
\VAR{\tilde{\theta}^t - \theta^t} 
\xrightarrow{ t  \rightarrow + \infty}    \frac{ [\frac{1}{N^2} + \frac{1}{(N-M_K)^2}] \sum_{j\in J} \left( M_j\rho_j' \right)^2 }{1- \left( \epsilon' + \frac{M_K}{N} \right)^{2}}
\end{equation}
The asymptotic variance still strictly increases with $M_K$. 

\underline{Note:} We introduce $x = \lambda (r_j + \mu) \frac{E M_j}{S}$. By taking the partial derivative of $\rho_j'$ with respect to $\mu$, we get:
\begin{align}
\frac{\delta\rho_j'}{\delta\mu}
= \frac{\lambda}{2S}\sigma_j^2\frac{1}{(r_j+\mu)^2} [-1 + (1 + 2x)e^{-2x}] ,
\end{align}
which is strictly negative for a positive $\mu$ considering that all the other constants are positive. Hence, the SGD noise variance $\rho_j'$ is inversely proportional with the regularization factor $\mu$.

Similarly, for $\epsilon'$, by considering that $\eta_j'$ can be rewritten as $\eta_j'= \gamma_j \frac{r_j}{r_j + \mu}  + \frac{\mu}{r_j + \mu}$, the partial derivative of $\eta_j'$ with respect to $\mu$ can be expressed as:
\begin{align}
\frac{\delta\eta_j'}{\delta\mu}
= \frac{r_j}{(r_j + \mu)^2}[1 - (1 - x) e^{-x}] ,
\end{align}
which is strictly positive. Hence $\eta_j'$ is strictly increasing with the regularization $\mu$ and so is $\epsilon'$.

Considering the behaviours of $\epsilon'$ and $\rho_j'$ with respect to the regularization term $\mu$, the more regularization is asked by the server and the smaller the asymptotic variance is, leading to more accurate free-riding attacks.

\textbf{Theorem \ref{theo:noise_additive} for FedProx}

The free-riders mimic the behaviour of the fair clients. Hence, we get:
\begin{align}
    {{\varphi_k}'}^2 = \frac{\lambda}{S}{\sigma_k}^2\frac{1}{2(r_j+\mu)}\left[1-e^{-2\lambda (r_k+ \mu )\frac{EM_j}{S}}\right]
\end{align}
leading to
\begin{align}
\VAR{\tilde{\theta}^t - \theta^t} 
&\xrightarrow{ t \rightarrow +\infty}    \frac{ [\frac{1}{N^2} + \frac{1}{(N-M_K)^2}] \sum_{j\in J} \left( M_j\rho_j'\right)^2 }{1- \left( \epsilon' + \frac{M_K}{N} \right)^{2}} \nonumber \\
& + \frac{1  }{1 - \left( \epsilon' + \frac{M_K}{N} \right)^{ 2 } } \frac{M_K^2}{N^2}\varphi'^2  .
\end{align}

For disguised free-riders, the variance is also inversely proportional to the regularization parameter $\mu$.

\textbf{Corollary \ref{cor:noise_additive_multi} for FedProx}

Similarly, for many free-riders, we get:
\begin{align}
\VAR{\tilde{\theta}^t - \theta^t} 
&\xrightarrow{ t \rightarrow +\infty}    \frac{ [\frac{1}{N^2} + \frac{1}{(N-M_K)^2}] \sum_{j\in J} \left( M_j\rho_j'\right)^2 }{1- \left( \epsilon' + \frac{M_K}{N} \right)^{2}} \nonumber \\
& + \frac{1  }{1 - \left( \epsilon' + \frac{M_K}{N} \right)^{ 2 } } \frac{M_K^2}{N^2}\sum_{k\in K} \varphi_k'^2  .
\end{align}

\section{Additional experimental results}
\label{app:plots}
\input{tex/extra_plots.tex}

%% file: tex/extra_plots.tex
\subsection{Accuracy Performances}
\label{supp:accuracy}

\begin{figure*}
    \centering
    \includegraphics[width=\linewidth]{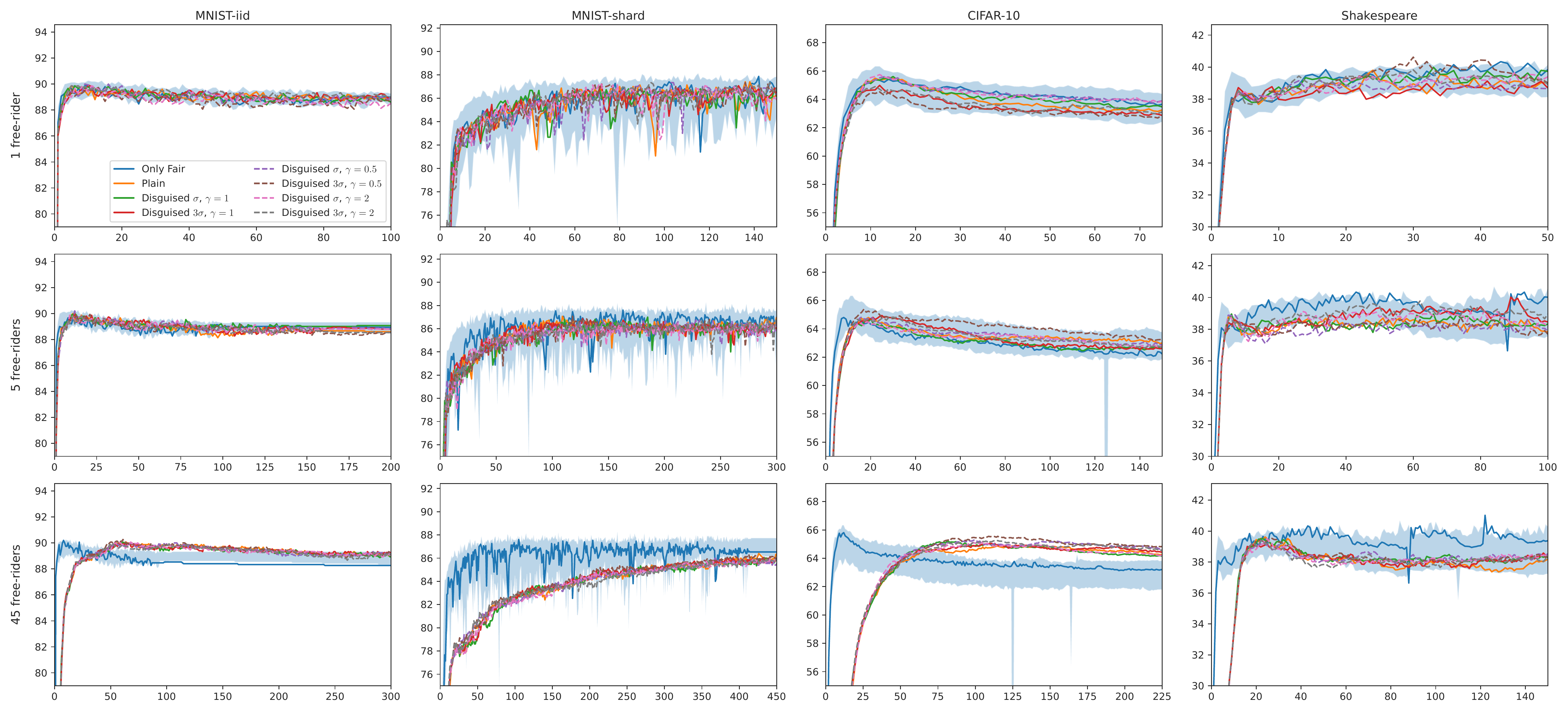}
    \caption{Accuracy performances for FedAvg and 20 epochs in the different experimental scenarios.}
\end{figure*}

\begin{figure*}
    \centering
    \includegraphics[width=\linewidth]{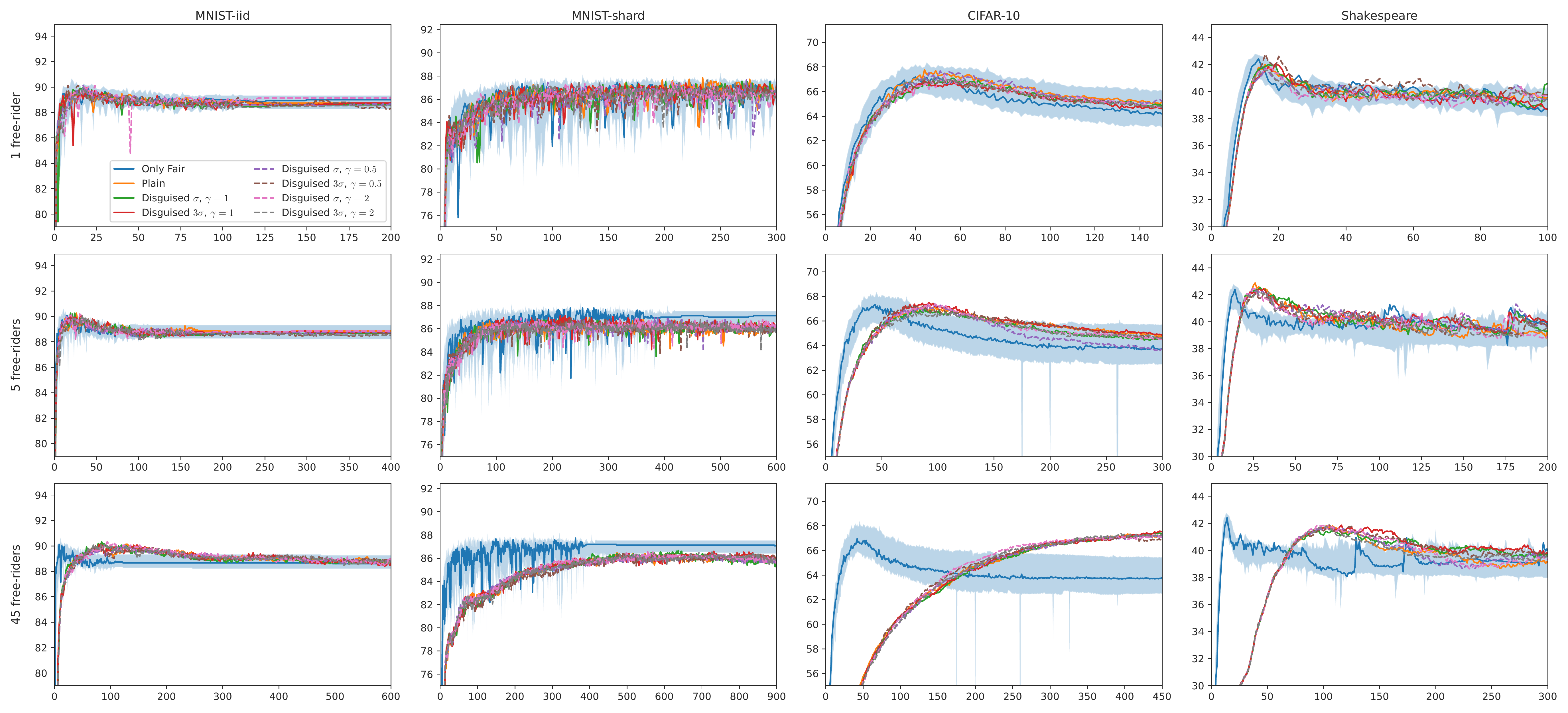}
    \caption{Accuracy performances for FedAvg and 5 epochs in the different experimental scenarios.}
\end{figure*}

\begin{figure*}
    \centering
    \includegraphics[width=\linewidth]{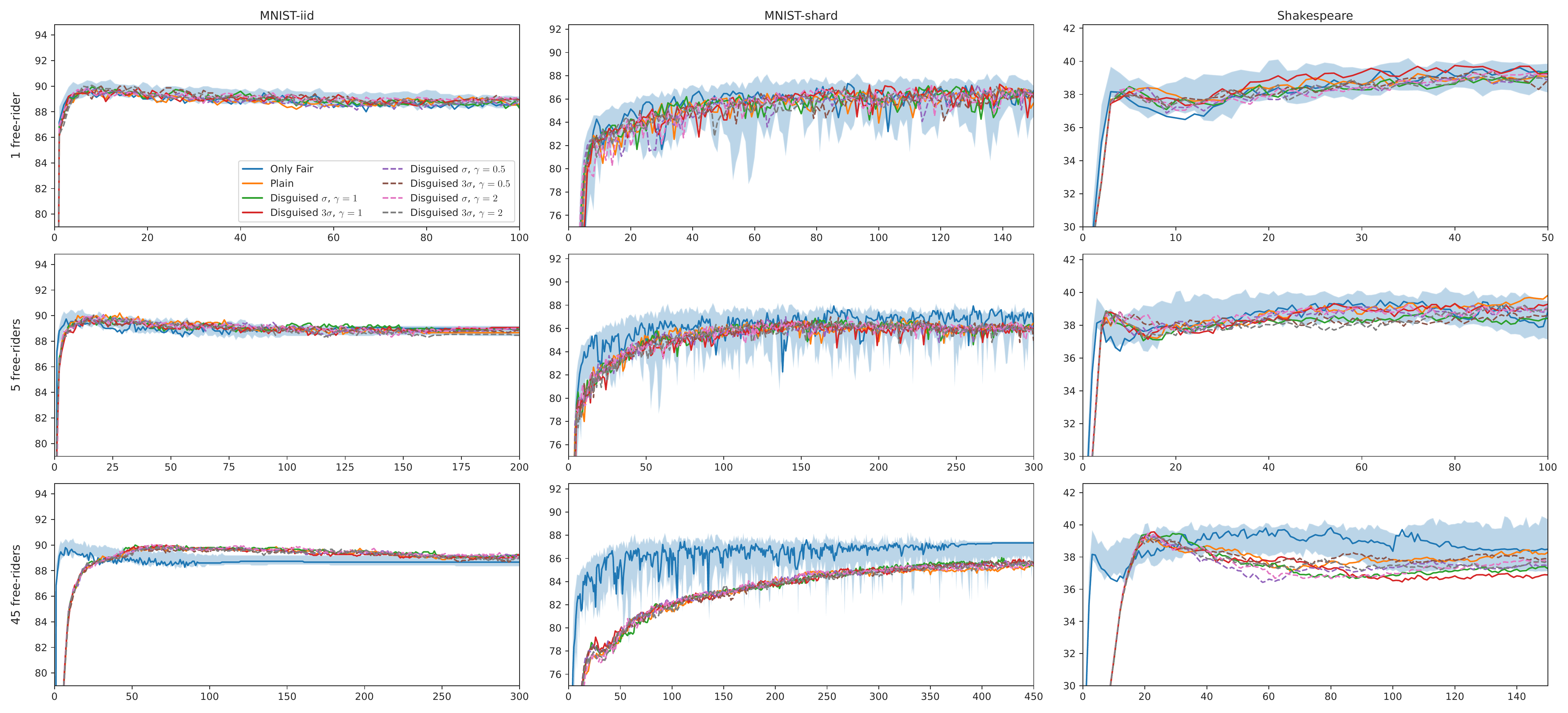}
    \caption{Accuracy performances for FedProx and 20 epochs in the different experimental scenarios.}
\end{figure*}

\begin{figure*}
    \centering
    \includegraphics[width=\linewidth]{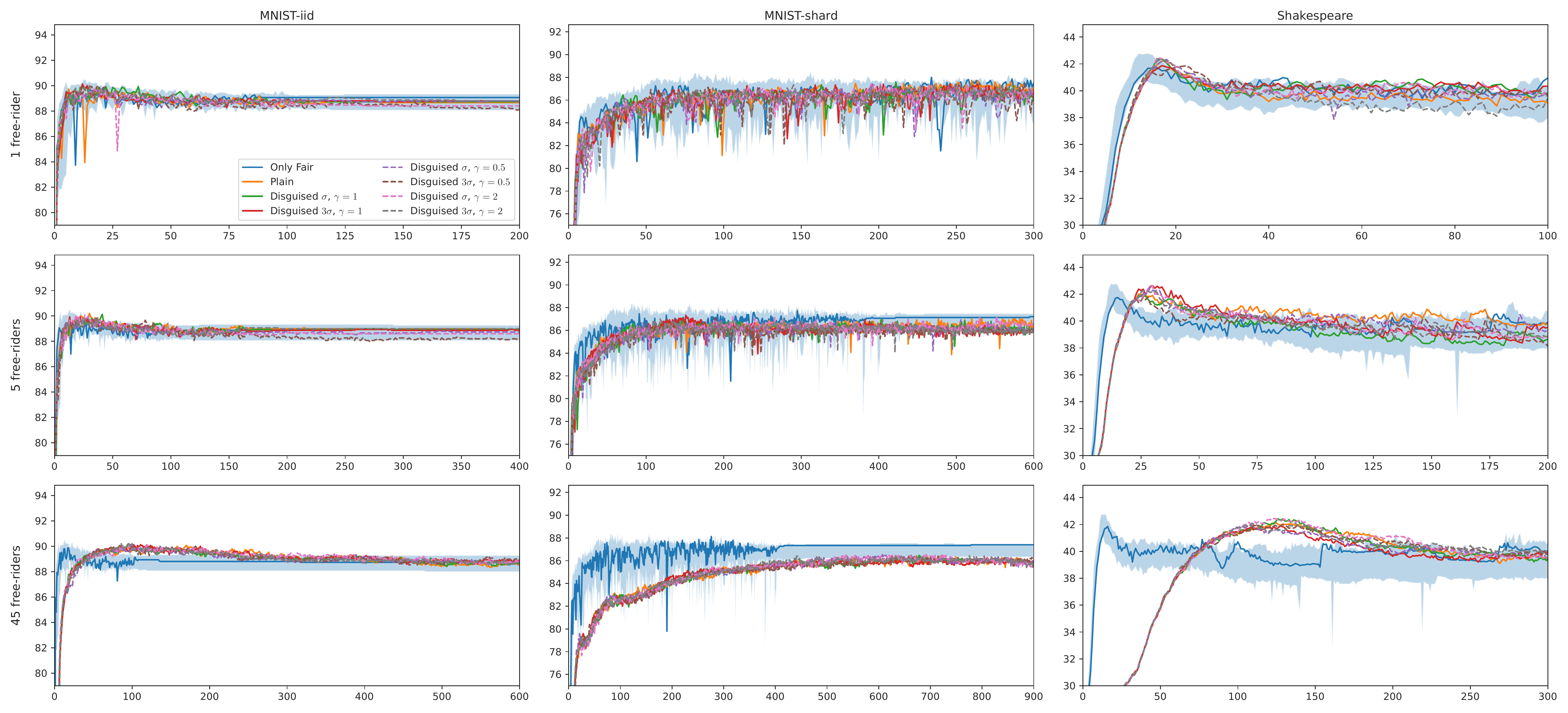}
    \caption{Accuracy performances for FedProx and 5 epochs in the different experimental scenarios.}
\end{figure*}

\begin{figure*}
    \centering
    \includegraphics[width=\linewidth]{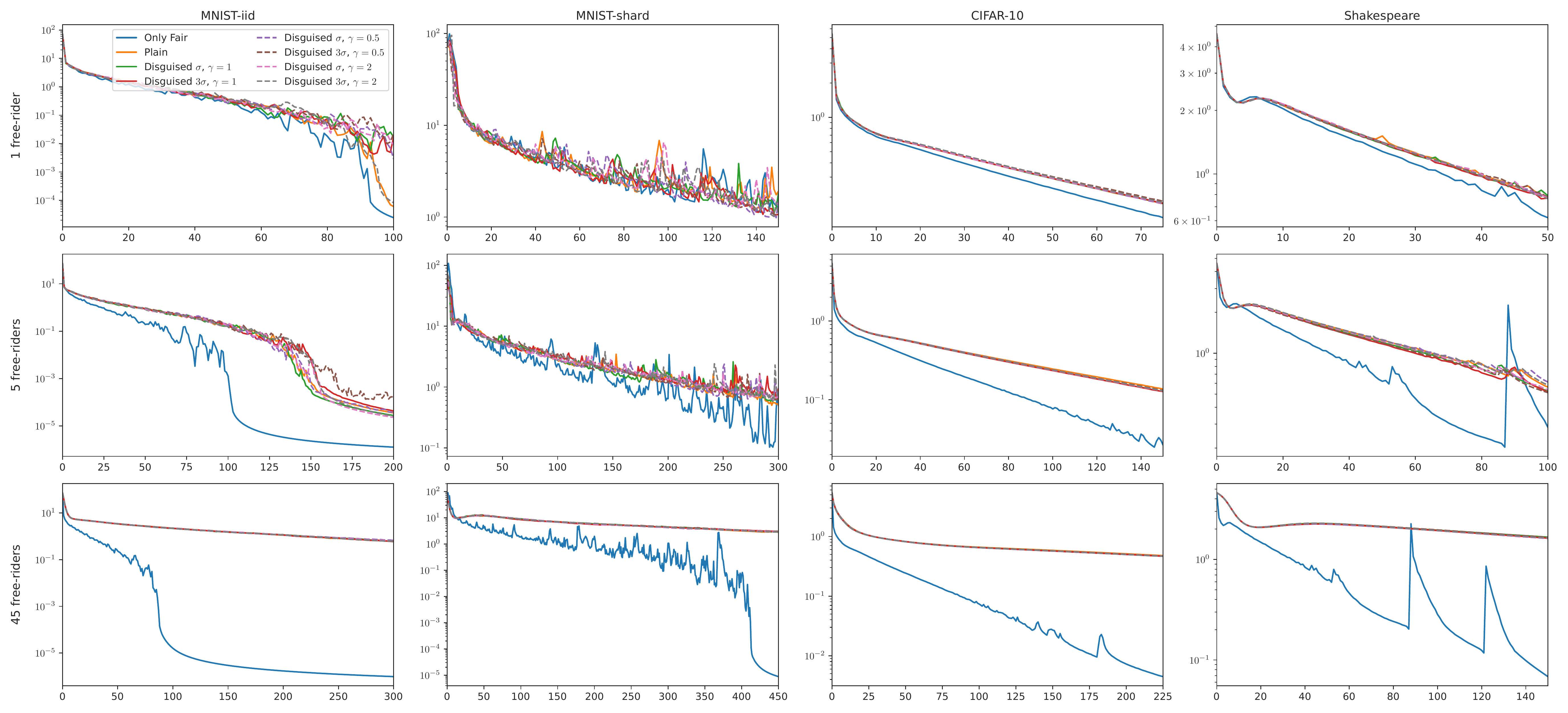}
    \caption{Loss performances for FedAvg and 20 epochs in the different experimental scenarios.}
\end{figure*}

\begin{figure*}
    \centering
    \includegraphics[width=\linewidth]{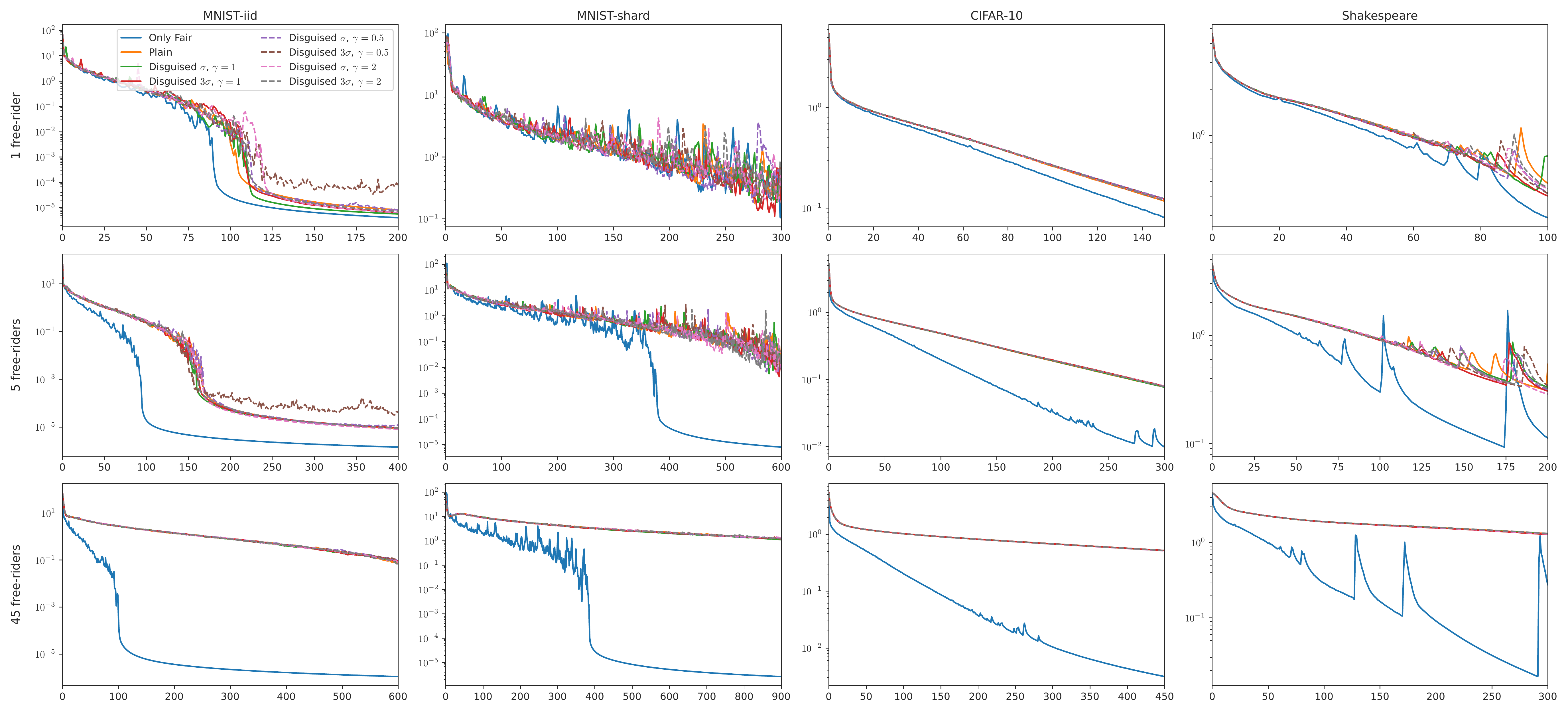}
    \caption{Loss performances for FedAvg and 5 epochs in the different experimental scenarios.}
\end{figure*}

\begin{figure*}
    \centering
    \includegraphics[width=\linewidth]{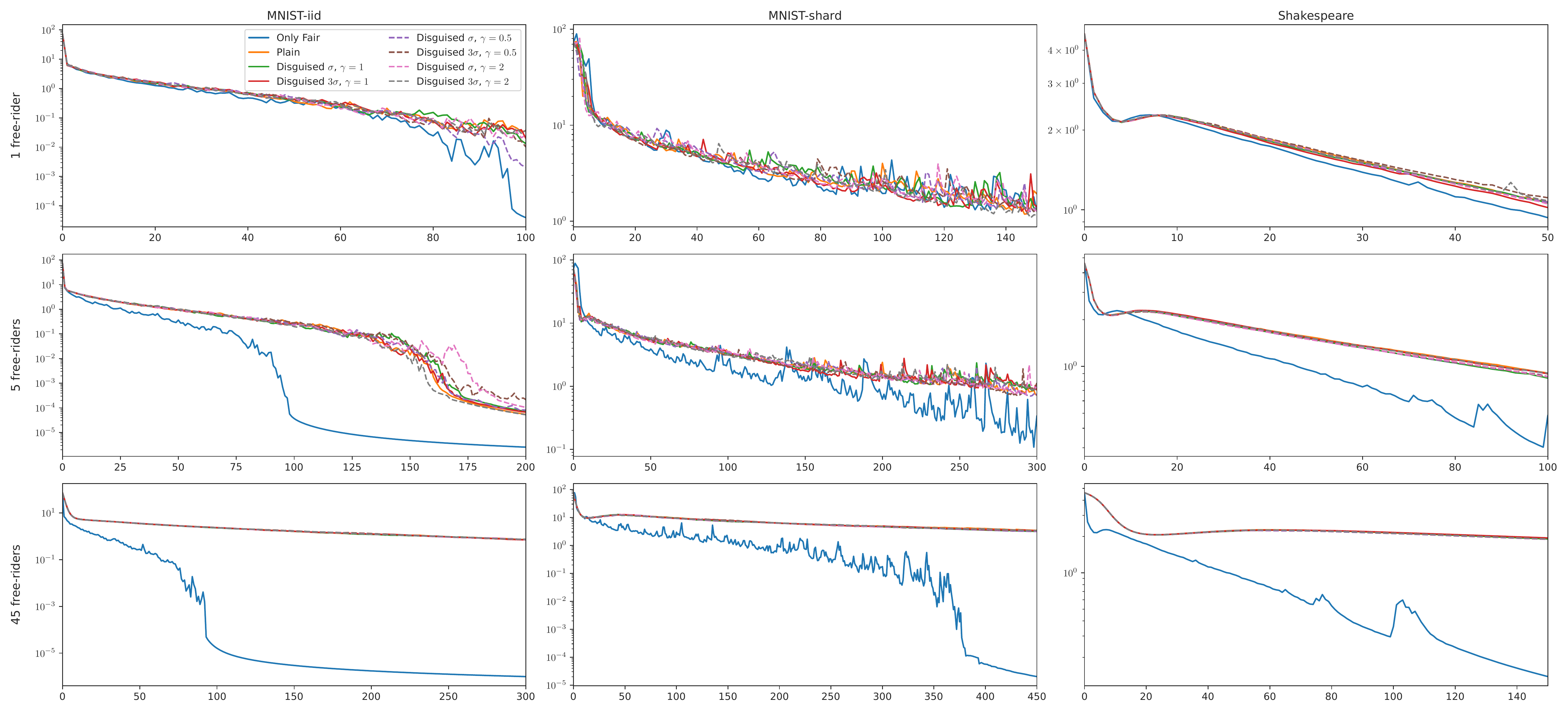}
    \caption{Loss performances for FedProx and 20 epochs in the different experimental scenarios.}
\end{figure*}

\begin{figure*}
    \centering
    \includegraphics[width=\linewidth]{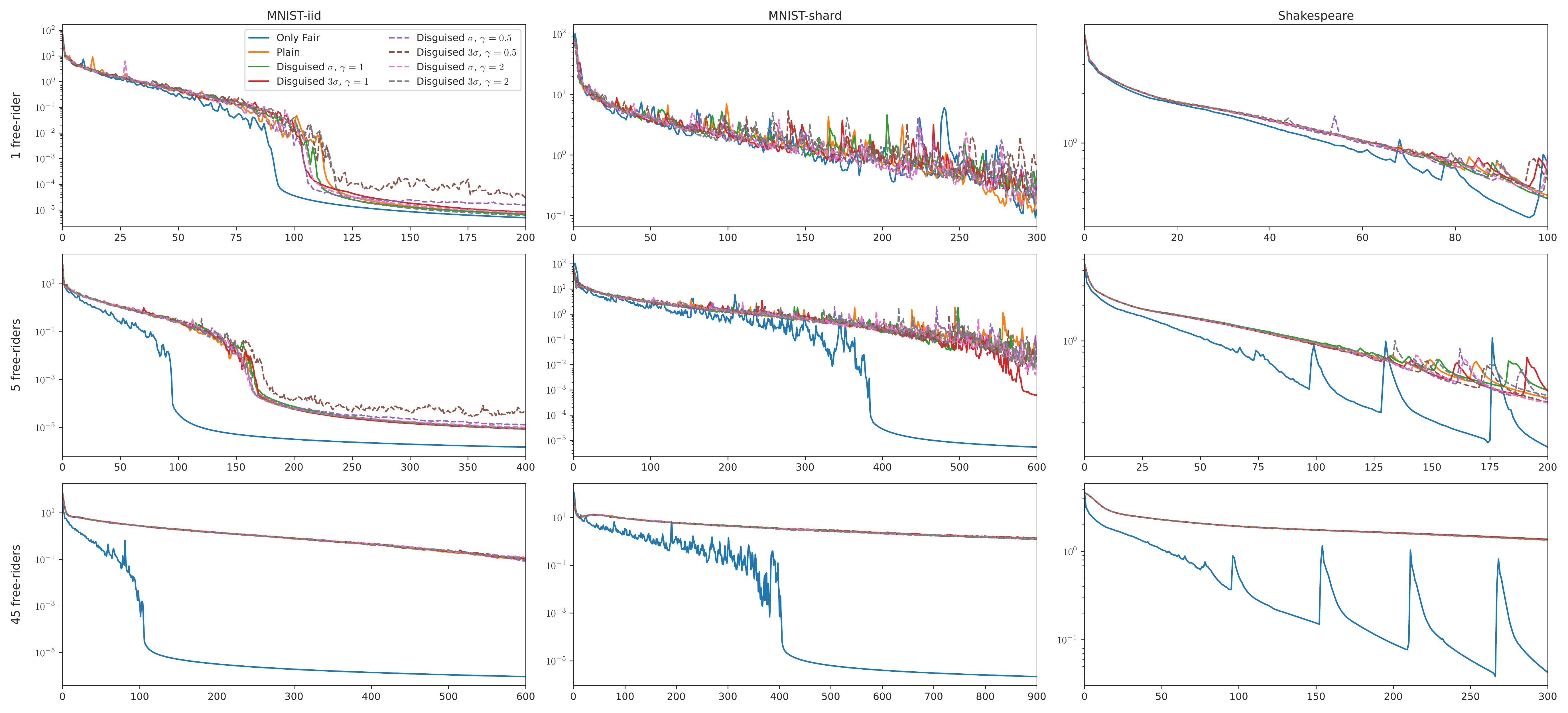}
    \caption{Loss performances for FedProx and 5 epochs in the different experimental scenarios.}
\end{figure*}